%% file: biasflow_main.tex
\documentclass[10pt,journal]{IEEEtran}

\usepackage{verbatim}
\usepackage{amsmath}
\usepackage{amsthm}
\usepackage{amssymb}
\usepackage{amsfonts}
\usepackage{graphicx}
\usepackage{url}
\usepackage{subfigure}
\usepackage{float}
\usepackage{mathrsfs,fourier,pifont}
\usepackage{threeparttable}
\usepackage{algorithm}
\usepackage{algpseudocode}
\usepackage{color}
\usepackage{xcolor}
\usepackage{colortbl}
\usepackage{booktabs}
\usepackage{epstopdf}
\usepackage{multicol,multienum, multirow}
\usepackage{makecell}

\usepackage{tablefootnote}
\usepackage{hyperref}
\usepackage{microtype}         
\usepackage{pgfplots}

\usepgfplotslibrary{groupplots}
\usetikzlibrary{patterns}
\usepgfplotslibrary{polar}
\usepackage{pgfplotstable}
\usepackage{siunitx}
\usepackage{tikz}
\usetikzlibrary{arrows}

\makeatletter
\def\th@plain{%
  \rm 
}
\makeatother


\newcommand{\colortext}[2]{#2}
\usepackage{pdflscape}
\newcommand{\suppincludegraphics}[2][]{\includegraphics[#1]{#2}}
\hypersetup{hidelinks,pdftitle={BiasFlow: Geometric Monitoring and Backbone Regularization for Spurious Feature Reliance},pdfauthor={Haojin Deng, Zhiping Lin, Yimin Yang},pdfsubject={Preprint}}

\begin{document}
\title{BiasFlow: Geometric Monitoring and Backbone Regularization for Spurious Feature Reliance}

\author{\colortext{blue}{Haojin~Deng, Zhiping~Lin, and Yimin~Yang}
\thanks{This work was supported in part by the Natural Sciences and Engineering Research Council of Canada (NSERC) Discovery Grant Program under Grant RGPIN-2025-06527, in part by the NSERC Alliance Program under Grant ALLRP 583061-23, and in part by the Western Research Award.} \thanks{H. Deng is with the Department of Electrical and Computer Engineering, Western University, London, Canada.} \thanks{Z. Lin is with the School of Electrical and Electronic Engineering, Nanyang Technological University, Singapore.} \thanks{Y. Yang is with the Department of Electrical and Computer Engineering, Western University, London, Canada, and also with the Vector Institute, Toronto, Canada.}}

\markboth{Preprint, October 2026}%
{Deng \MakeLowercase{\textit{et al.}}: BiasFlow: Geometric Monitoring and Backbone Regularization}
\maketitle

\begin{abstract}
\colortext{blue}{Worst-group accuracy (WGA) evaluates a trained predictor but does not characterize how its frozen backbone behaves when a new head is learned. We introduce \textbf{BiasFlow}, a hook-based toolkit for monitoring class--attribute centroid alignment (IBMI), within-class centroid separation (W-IBMI), and feature-projection sensitivity. IBMI is confounded by class--attribute correlation and is not a measure of causal feature reliance. We pair these diagnostics with \textbf{BiasFlow Regularization (BFR)}, a supervised, composable class-conditional centroid-alignment penalty. W-IBMI verifies the quantity BFR optimizes; it is scale dependent and does not independently establish attribute removal. Across the reported small-scale benchmarks, adding BFR improves or preserves mean WGA, with gains up to $+26.0$\,pp on UrbanCars. The principal independent stress test freezes CelebA-Std backbones and trains fresh heads on biased data: BFR+GroupDRO improves WGA from $40.7\%$ to $64.1\%$, while Male probe accuracy decreases from $92.5\%$ to $72.2\%$. Attribute information remains recoverable, and cross-task results are mixed. A controlled synthetic-watermark ImageNet experiment additionally improves watermark-shift accuracy by $+23.0$\,pp under matched training. These results support evaluating centroid geometry and resistance to biased head retraining alongside WGA, within the tested protocols.}

\end{abstract}


\section{Introduction}
\label{sec:introduction}

\colortext{blue}{Spurious correlations can arise when a target label co-occurs with a task-irrelevant attribute in the training data. Empirical Risk Minimization (ERM) may exploit such correlations and fail when they change~\cite{geirhos2020shortcut}. This problem occurs in facial attribute recognition~\cite{liu2015celeba} and visual group-shift benchmarks~\cite{sagawa2020groupdro}.}

\begin{figure}[t]
\centering
\includegraphics[width=\columnwidth]{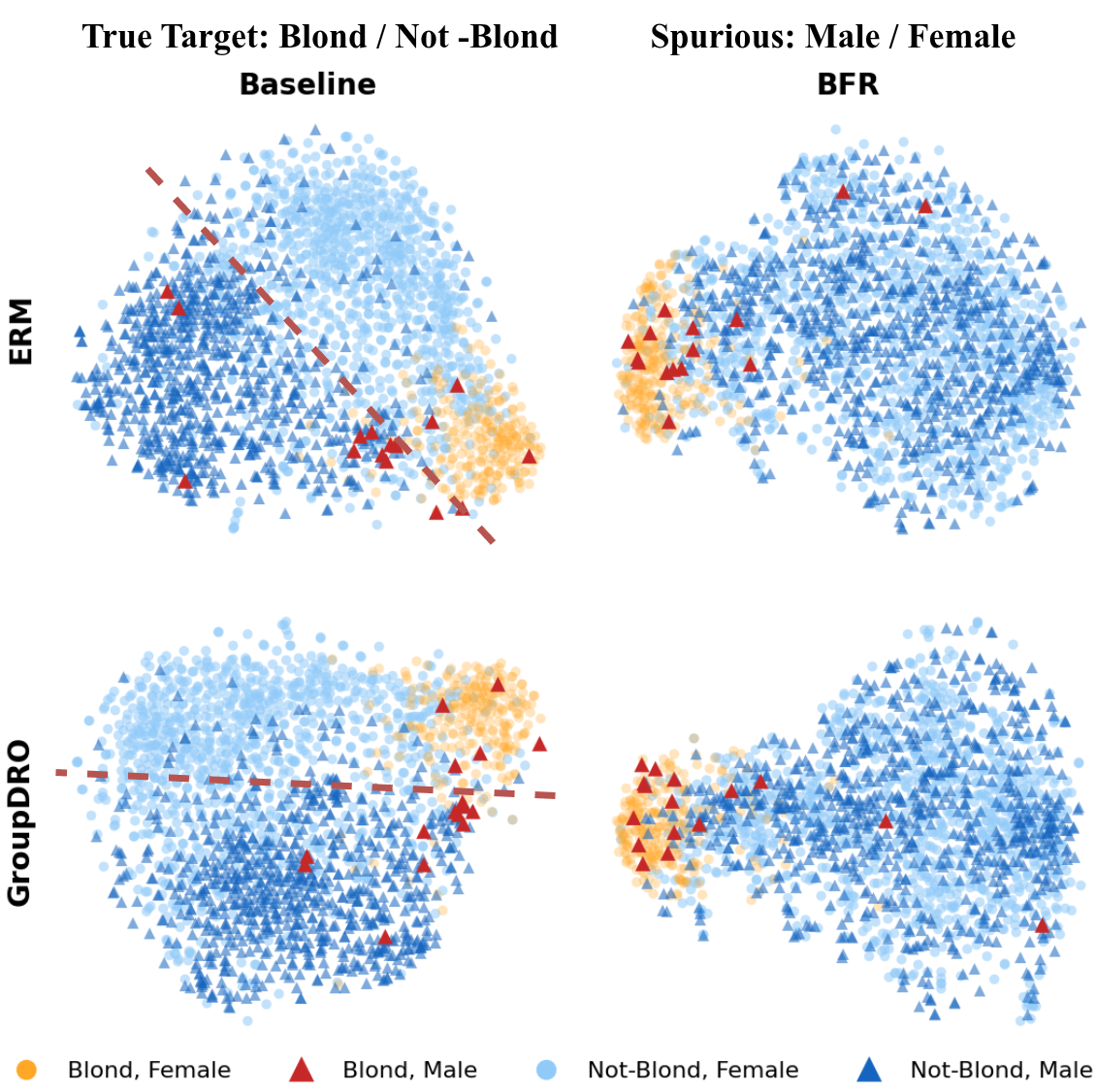}
\caption{\colortext{blue}{t-SNE of CelebA-Std ResNet-18 features, seed 1 (Blond Hair target, Male attribute). Baseline and BFR panels illustrate different within-class subgroup layouts. BFR reduces the targeted centroid discrepancy; two-dimensional mixing does not establish concept erasure or causal feature use.}}
\label{fig:tsne}
\end{figure}

Over the past several years, the community has developed a rich family of debiasing methods aimed at minimizing spurious reliance while preserving overall model performance. Loss-reweighting strategies such as LfF~\cite{nam2020lff} and JTT~\cite{liu2021jtt}, head-retraining approaches such as DFR~\cite{kirichenko2023dfr}, adversarial discoverers such as DebiAN~\cite{li2022debian}, and group-robust optimizers such as GroupDRO~\cite{sagawa2020groupdro} have all reported substantial improvements on the standard evaluation benchmark: Worst-Group Accuracy (WGA), \colortext{blue}{the minimum accuracy over evaluated groups defined by the cross-product of class and spurious attribute.}

\colortext{blue}{WGA is the minimum accuracy over predefined evaluation groups; it does not summarize confidence margins, representation geometry, or the behavior of a subsequently trained classifier. Models with different representations can have the same WGA. DFR~\cite{kirichenko2023dfr} demonstrates that a group-balanced head can improve WGA while its input backbone is held fixed. This is a useful strategy, not evidence that the original classifier necessarily uses every attribute encoded by its backbone. We ask the narrower question of whether a frozen representation remains susceptible to a new head trained under the original biased distribution, consistent with the distinction between representation and classifier studied in~\cite{kang2019decoupling,geirhos2020shortcut}.}

\colortext{blue}{We introduce \textbf{BiasFlow} to inspect this distinction through geometric measurements and head-retraining stress tests. The \textbf{Inter-class Bias-alignment Metric Index (IBMI)} measures centroid alignment with a pooled, user-specified attribute direction. Because class composition contributes to that direction, a high IBMI can occur even without within-class attribute separation. We therefore interpret IBMI as an observational summary, not proof of reliance or contamination. DFR preserves its own first-stage backbone during head retraining; equality with a separately trained ERM run requires identical checkpoint and probe conditions, which the audit does not enforce. The projection probe defines signed group accuracy changes, but the retained figure plots unsigned AblGap. Its nonnegative bars establish neither exploitation nor interference.}

\colortext{blue}{The independent stress test is particularly informative on CelebA-Std\footnote{CelebA-Std uses Blond Hair as target and Male as the spurious attribute, with the Blond-Male group rare~\cite{sagawa2020groupdro}.}. GroupDRO achieves $86.3\%$ end-to-end WGA, but a fresh head trained on the standard biased data over its frozen backbone obtains $40.7\%$, close to ERM's $40.0\%$ (Table~\ref{tab:exp2_probe}). Male probe accuracy is $92.5\%$. These are distinct observations: a new biased head performs poorly, and a separate probe can recover the attribute. Neither establishes that GroupDRO's original head uses Male causally.}

\colortext{blue}{Our intervention is deliberately restricted to the first moments of class-conditional attribute groups. \textbf{BiasFlow Regularization (BFR)} penalizes the distance between their feature centroids during backbone training. It belongs to the established family of conditional representation-alignment methods~\cite{zhao2020conditional,zhang2022cnc,park2026scer}; its role here is a simple composable intervention paired with geometric monitoring and a frozen-backbone evaluation. Equal centroids do not imply equal feature distributions or attribute erasure.}

\colortext{blue}{BFR reduces the objective-aligned W-IBMI statistic in most configurations (Table~\ref{tab:bfr_wibmi}). More independently, BFR+GroupDRO yields $64.1\%$ biased-head WGA and $72.2\%$ Male probe accuracy, compared with $40.7\%$ and $92.5\%$ without BFR. The former tests resistance to biased head retraining; the latter measures recoverability under the specified probe. Figure~\ref{fig:tsne} illustrates within-class subgroup mixing in a two-dimensional visualization, which should not be interpreted as proof of complete distribution alignment.}

Our contributions are summarized as follows:

\begin{itemize}
\colortext{blue}{\item \textbf{Geometric monitoring with explicit scope.} BiasFlow reports pooled centroid alignment, class-conditional centroid separation, and projection sensitivity. We specify class-composition and scale limitations, and distinguish the unsigned AblGap from signed group accuracy drops. These summaries complement WGA; they do not certify fairness, causal reliance, or concept removal.}

\colortext{blue}{\item \textbf{Composable supervised centroid regularization.} BFR applies class-conditional centroid alignment to the evaluated training objectives. We retain the reported WGA results, including the near-neutral Waterbirds JTT case, and distinguish supervised comparisons from gains over methods without training-time attribute labels. We do not claim that conditional alignment itself is new or that BFR outperforms unevaluated representation-level methods.}

\colortext{blue}{\item \textbf{Frozen-backbone stress testing.} Fresh biased heads on CelebA-Std provide the primary independent evidence: GroupDRO's backbone supports $40.7\%$ WGA, versus $64.1\%$ for BFR+GroupDRO. Appendix experiments report continued ERM training, Smiling transfer, and alternative attribute probes, including negative outcomes. A separate controlled synthetic-watermark study tests a large-scale nuisance intervention under matched training.}

\end{itemize}

\section{Related Work}
\label{sec:related_work}

\colortext{blue}{Spurious-correlation mitigation includes predictive-loss methods, head retraining, and a substantial literature on representation alignment, adversarial removal, and concept erasure. We discuss these families to position BFR as a centroid-level instance of representation regularization rather than the first representation-level intervention.}

\colortext{blue}{\subsection{Predictive-Loss and Head-Retraining Methods}}

\colortext{blue}{The baselines evaluated here span sample reweighting, head retraining, and group-supervised robust optimization. Reweighting and group-loss training can change the entire network, including the backbone; only the head-retraining stage of DFR explicitly leaves the backbone fixed.}

\textbf{Loss-Reweighting Approaches.} These methods aim to improve Worst-Group Accuracy (WGA) by dynamically altering the importance of training samples without requiring explicit group annotations. Just Train Twice (JTT) \cite{liu2021jtt} identifies misclassified samples from a standard ERM model and upweights them in a second training phase. Learning from Failure (LfF) \cite{nam2020lff} simultaneously trains a biased network to amplify prejudice and a debiased network that focuses on samples the biased model struggles with. Similarly, Debiasing Alternate Networks (DebiAN) \cite{li2022debian} employs an adversarial setup to discover and unlearn unknown biases. These methods can shift the final decision rule toward minority or shortcut-conflicting samples, but they regulate training through sample weights rather than explicit representation constraints. Consequently, the backbone remains free to encode spurious attributes even when the output classifier becomes more robust.

\colortext{blue}{\textbf{Classification Head Retraining.} DFR~\cite{kirichenko2023dfr} retrains a linear head on balanced data over a frozen ERM-trained backbone. It demonstrates that useful core features can already be present. Our stress test concerns a later biased head, not a defect in the balanced DFR decision rule.}

\colortext{blue}{Head retraining preserves an existing representation and can avoid the cost of relearning it. BFR instead changes the backbone during its training stage. These interventions answer different questions: robust prediction with a selected head, versus the behavior of new heads on frozen features. Neither approach universally dominates, particularly when transferred tasks benefit from the targeted attribute.}

\textbf{Debiasing with Explicit Bias Annotations.} When spurious group labels are available, methods can directly enforce robustness. Invariant Risk Minimization (IRM)~\cite{arjovsky2019coloredmnist} seeks representations that support invariant prediction across training environments, offering a theoretical route to avoiding environment-specific shortcuts. Group Distributionally Robust Optimization (GroupDRO) \cite{sagawa2020groupdro} minimizes the maximum loss across predefined groups. Recent advancements have extended this paradigm: Hierarchical Group DRO \cite{hierarchicalDRO2025} addresses both inter-group and intra-group distributional shifts through a multi-level ambiguity set, while LISA (Learning Invariant Predictors with Selective Augmentation) \cite{yao2022lisa} leverages group labels to perform selective mixup interpolation, effectively canceling out spurious correlations during training. These methods enforce outcome robustness through group-level losses or augmentations, but high WGA does not by itself certify that the backbone has become invariant to the spurious attribute.


\begin{figure*}[!ht]
	\centering
	\resizebox{\textwidth}{!}{\input{plots/framework.tex}}
	\caption{\colortext{blue}{BiasFlow workflow and evidence roles. (a) Forward hooks expose centroid geometry and unsigned projection sensitivity. Blue circles and orange triangles denote attribute values, dashed ellipses denote target classes, and open markers denote group centroids. (b) BFR-inv penalizes within-class centroid differences, averaged over annotated attributes during backbone training; W-IBMI is scale-dependent, objective-aligned verification. (c) Frozen-backbone tests fit a fresh task head on biased training features and a separate attribute probe on validation features, then evaluate on test features. Geometry is schematic, not measured data. Centroid alignment and projection sensitivity do not establish concept erasure or a causal intervention-selection rule; predictive evidence is limited to the tested protocols.}}
	\label{fig:framework}
\end{figure*}
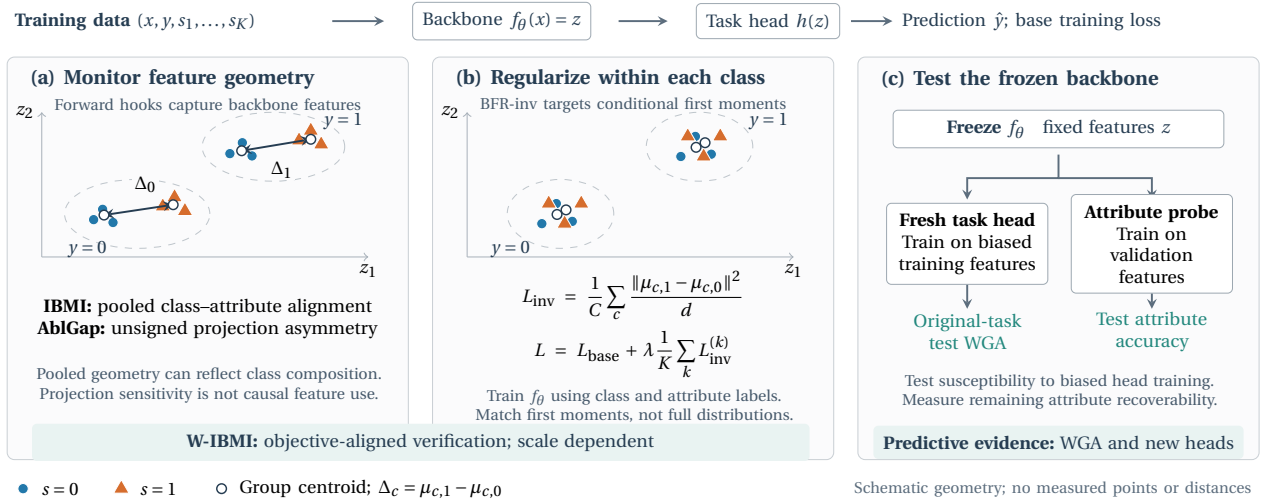

\subsection{Latent and Causal Fairness Metrics}
To look beyond output accuracy, recent work has developed metrics to quantify bias directly within learned features.

\textbf{Observational Probes.} Information-theoretic approaches measure statistical dependence. For instance, InfoFair \cite{kang2022infofair} minimizes variational mutual information between outcomes and sensitive attributes, while Gowri et al. \cite{gowri2024approximating} propose scalable approximations for high-dimensional latent mutual information. Geometrically, Zhou et al. \cite{zhou2023faircca} introduce Fair CCA to minimize correlation disparity across groups via canonical correlations, and Frolov and Oseledets \cite{frolov2019quantifying} use principal angle alignments to quantify subspace relationships. These metrics expose structure that output accuracy cannot see, but they are generally not designed as lightweight monitoring tools for debiasing pipelines or tied to a corresponding backbone-level training objective.

\textbf{Interventional Probing.} Interventional metrics test whether representations actively influence predictions rather than merely encoding information. Guerner et al. \cite{guerner2023geometric} perform feature-space interventions on learned subspaces to verify if specific concepts drive model predictions, offering stronger evidence of debiasing than observational correlation. However, such probes are usually used as post-hoc analyses; they rarely distinguish whether a spurious feature is being exploited as a shortcut or acting as interference, and they do not prescribe how to reshape the backbone.

\subsection{Relationship to Prior Work}
\colortext{blue}{Conditional alignment is established prior work. \emph{Conditional Learning of Fair Representations}~\cite{zhao2020conditional} aligns representations across demographic groups conditional on class to address equalized error rates. \emph{Correct-N-Contrast} (CNC)~\cite{zhang2022cnc} uses an ERM model to identify same-class examples with different spurious cues and learns aligned representations contrastively without training-time attribute labels; it also connects alignment to worst-group error. BFR uses supplied attribute labels and a first-moment penalty, so it offers neither CNC's annotation advantage nor full conditional distribution matching.}

\colortext{blue}{\emph{Learning Not to Learn}~\cite{kim2019learningnot} adversarially suppresses bias information through a feature--bias mutual-information objective. \emph{EnD}~\cite{tartaglione2021end} regularizes intermediate representations by entangling same-target features and disentangling bias-related features. \emph{LEACE}~\cite{belrose2023leace} performs closed-form linear concept erasure with guarantees for the transformed representation; BFR's approximate conditional centroid alignment has no comparable erasure guarantee. \emph{Spurious Correlation-Aware Embedding Regularization} (SCER)~\cite{park2026scer} derives core and spurious directions from group-wise mean embeddings and regularizes the embedding space for worst-group robustness. SCER is especially close in its use of group means, while its direction- and classifier-based theoretical motivation differs from BFR's direct within-class centroid penalty. Our contribution is the monitoring and evaluation workflow around a composable penalty. The experiments do not include these close representation-level baselines and cannot establish superiority over them.}

\section{Methodology: BiasFlow and Representation Regularization}
\label{sec:methodology}

\colortext{blue}{BiasFlow collects latent features with forward hooks to compute observational geometry and feature-projection sensitivity without changing baseline training. BFR is a separate supervised training penalty applied to the features. Figure~\ref{fig:framework} summarizes the workflow. The metric limitations below apply to all models, including ordinary ERM backbones, rather than arising only after BFR.}

\noindent\textbf{Notation.} Let $f_\theta(x)\in\mathbb{R}^d$ denote the backbone feature embedding of input $x$, $y\in\{1,\dots,C\}$ the class label, $s\in\{0,1\}$ the binary spurious attribute, $\mu_c$ the empirical centroid of class $c$, and $S_B$ the between-class scatter matrix. These symbols are shared across both components below.

\subsection{BiasFlow Diagnostic: IBMI-based Geometric Probing}
\label{subsec:biasflow_probe}

The first component of BiasFlow instruments any backbone via forward hooks to compute a hierarchy of geometric metrics that quantify spurious entanglement at the representation level, without modifying training.

\noindent\textbf{Geometric foundations.}
We define $\mu_c$ as the empirical class centroid and $w_{\mathrm{cls}}$ as the principal eigenvector of the between-class scatter matrix $S_B$, following Fisher~\cite{fisher1936lda}. The spurious direction $w_{\mathrm{spur}}$ is the unit vector connecting the two spurious-group centroids:
\begin{equation}
\label{eq:spur-axis}
w_{\mathrm{spur}} \;=\; \frac{\mu(s=1) - \mu(s=0)}{\|\mu(s=1) - \mu(s=0)\|_2}\,.
\end{equation}
\colortext{blue}{Here $s$ is a user-specified attribute such as Male, background, a co-occurring object, or a synthetic watermark. The pooled mean satisfies $\mu(s)=\sum_c P(y=c\mid s)\mu(y=c,s)$. Thus $w_{\mathrm{spur}}$ can contain class-composition differences even when $\mu(y=c,s=1)=\mu(y=c,s=0)$ for every class. We call it an attribute-associated direction rather than a causally isolated spurious factor. It is recomputed per model, so cross-model differences also involve changes in the probe axis. If the centroid difference vanishes, the direction is undefined and the corresponding statistic should be marked unavailable.}

\noindent\textbf{Inter-class Bias-alignment Metric Index (IBMI).}
\colortext{blue}{IBMI measures the variance of equally weighted class centroids projected on $w_{\mathrm{spur}}$, relative to the largest eigenvalue of their scatter. For $C>1$, let $\bar\mu=C^{-1}\sum_c\mu_c$ and $S_B=(C-1)^{-1}\sum_c(\mu_c-\bar\mu)(\mu_c-\bar\mu)^\top$. Use the same sample-variance normalization in the numerator:}
\begin{equation}
\label{eq:ibmi}
\mathrm{IBMI} \;=\; \frac{\operatorname{Var}_{c}\bigl(\langle \mu_c, w_{\mathrm{spur}}\rangle\bigr)}{\lambda_{\max}(S_B) + \varepsilon}\,,
\end{equation}
\colortext{blue}{where $\varepsilon>0$ stabilizes division. The numerator equals $w_{\mathrm{spur}}^\top S_Bw_{\mathrm{spur}}$, so the Rayleigh bound gives $0\leq\mathrm{IBMI}\leq1$ for a unit axis and positive-semidefinite $S_B$. Degenerate class scatter has no useful alignment interpretation and is treated as unavailable by the probe. For two noncoincident class centroids, IBMI is a squared cosine between the class- and attribute-centroid directions, up to $\varepsilon$. This is a standard geometric quantity, not a novel dependence estimator. High IBMI does not establish within-class attribute encoding or classifier use.}

\noindent\textbf{Within-Class IBMI (W-IBMI).}
\colortext{blue}{W-IBMI measures class-conditional centroid separation using the same supplied attribute labels as BFR. It avoids the pooled-axis confounding mechanism in its definition, but closely matches the BFR-inv objective. With class--attribute imbalance, successful centroid alignment can leave a pooled axis dominated by class composition, so IBMI may increase toward one despite the reduction in within-class separation. This is a limitation of interpreting pooled alignment as bias, not a reason to treat W-IBMI as independent validation:}
\begin{equation}
\label{eq:wibmi}
\mathrm{W\text{-}IBMI} \;=\; \frac{1}{C}\sum_{c=1}^{C} \frac{\bigl\|\mu(y{=}c,\,s{=}1) - \mu(y{=}c,\,s{=}0)\bigr\|^2}{d}\,,
\end{equation}
\colortext{blue}{where division by $d$ reports squared centroid distance per feature coordinate. It does not make the metric scale invariant or comparable across architectures: replacing $z$ by $\alpha z$ multiplies W-IBMI by $\alpha^2$, even if an inverse head-weight rescaling leaves predictions unchanged. We retain the reported values as objective-aligned verification statistics under each feature convention. They are not used to rank architectures or prove removal of attribute information.}

\colortext{blue}{IBMI and W-IBMI are descriptive first-moment summaries. Equal within-class centroids do not imply equal conditional distributions or prevent attribute prediction through other feature structure. Attribute probes and biased-head WGA therefore supply separate operational evidence. Balanced or class-residualized axes, normalized or whitened distances, feature norms, and distributional measures would be useful additional validations; no such new measurements are claimed here.}

\subsection{AblGap: Feature-Space Ablation Probe}
\label{subsec:ablgap}

IBMI and W-IBMI quantify the geometric presence of spurious structure in the backbone, but do not reveal the functional role that structure plays in the model's decisions. BiasFlow therefore includes a complementary feature-space ablation probe, the Ablation Gap (AblGap), which measures the accuracy consequence of projecting out the empirical prespecified-attribute direction from the feature space. Like IBMI, AblGap should be interpreted as a sensitivity test for a user-specified attribute direction, not as proof that this direction is a causally pure spurious factor.

\noindent\textbf{Definition.}
Given feature $z = f_\theta(x)$ and prespecified attribute direction $w_{\mathrm{spur}}$, we project out the corresponding component:
\begin{equation}
\label{eq:ablation}
\tilde{z} \;=\; z \;-\; \bigl\langle z,\, w_{\mathrm{spur}}\bigr\rangle\, w_{\mathrm{spur}}\,.
\end{equation}
Let $\mathcal{D}_\mathrm{R}$ and $\mathcal{D}_\mathrm{U}$ be the spurious-related and spurious-unrelated groups. The accuracy drop after projection for group $g \in \{\mathrm{R}, \mathrm{U}\}$ is:
\begin{equation}
\label{eq:acc-drop}
\Delta\mathrm{Acc}_g \;=\; \mathrm{Acc}_g(z) \;-\; \mathrm{Acc}_g(\tilde{z})\,,
\end{equation}
and AblGap $= |\Delta\mathrm{Acc}_R - \Delta\mathrm{Acc}_U|$ captures the asymmetry between groups.

\colortext{blue}{\noindent\textbf{Signed projection sensitivity.} AblGap is nonnegative and measures group asymmetry; it has no exploitation/interference sign. The sign belongs to $\Delta\mathrm{Acc}_R$: a positive value means the removed component helped the fixed head on group R, and a negative value means projection helped that group. We use exploitation-like and interference-like as tentative interpretations. Since the projected direction can include class information and projection creates a feature-space perturbation, neither sign isolates the underlying attribute causally. The retained CelebA and UrbanCars plot does not report the signed group drops; controlled validation of both sign interpretations and intervention selection remains future work. The probe is applied to trained baseline models before BFR intervention, not to an untrained model or to a dataset in isolation.}

\subsection{BiasFlow Regularization (BFR)}
\label{subsec:bfr}

\colortext{blue}{We test whether supervised within-class centroid alignment changes the behavior of the evaluated backbones. BFR adds a geometric penalty to the base training loss (cross-entropy in the ERM case):}
\begin{equation}
\label{eq:total-loss}
L_{\mathrm{total}} \;=\; L_{CE} \;+\; \lambda \cdot L_{\mathrm{geo}}\,,
\end{equation}
\colortext{blue}{where $\lambda$ is an effective scheduled weight. BFR-inv is the principal intervention; BFR-orth and BFR-spectral are additional modes evaluated in the appendix. The schedule and centroid handling are specified below. No validated exploitation/interference rule is used to choose a mode.}

\subsubsection{Within-Class Invariance Loss (BFR-inv)}
The \texttt{inv} mode directly targets the within-class component of spurious geometry. It aims to reduce the within-class spurious centroid signal by pulling the related ($s=1$) and unrelated ($s=0$) subsets of each class $c$ toward the same feature centroid:
\begin{equation}
\label{eq:inv-loss}
L_{\mathrm{inv}} \;=\; \frac{1}{C} \sum_{c=1}^C \frac{\|\mu(y=c, s=1) - \mu(y=c, s=0)\|^2}{d}\,,
\end{equation}
where $d$ is the feature dimension. \\
\colortext{blue}{\textbf{Mechanism and scope.} This penalty aligns first moments conditional on the target label. It does not impose independence of the attribute and representation, align all distributions, or guarantee erasure accessible to every classifier. We verify optimization with W-IBMI and evaluate independent predictive consequences using WGA and frozen-backbone stress tests.

\colortext{blue}{\textbf{Multiple attributes.} With $K$ annotated binary attributes, the implementation uses $L_{\mathrm{geo}}=K^{-1}\sum_{k=1}^K L_{\mathrm{inv}}^{(k)}$, each computed separately conditional on class and attribute $s_k$. CelebA-MB uses Male and Young, and UrbanCars uses background and co-occurring object. Both labels are required. This marginal alignment does not explicitly align all joint attribute combinations or constrain their interactions.}

\colortext{blue}{\textbf{Centroid estimates and gradients.} For each $(c,s_k)$ cell with at least two minibatch examples, the loss uses the current differentiable batch mean. A detached running buffer is initialized at the first qualifying batch and then updated as $m\leftarrow\alpha m+(1-\alpha)\hat\mu$, with $\alpha=0.9$. If a cell has fewer than two examples, its previous detached buffer is used; if no buffer exists, that class--attribute pair is skipped. The loss averages over classes with both centroids available and then over attributes. Gradients propagate only through current batch means. A term using two stored means has no backbone gradient, although it contributes to the reported scalar loss. Rare groups may receive delayed or intermittent alignment signals and stale estimates as the backbone evolves.}

\colortext{blue}{\textbf{Weight schedule.} For total training epochs $T$, let $e_w$ be the integer warmup length (fractional warmup overrides an absolute value), $r$ the configured ramp fraction, and $q$ an optional stop fraction. The effective weight is zero for $e\le e_w$ and, when enabled, for $e>\lfloor qT\rfloor$. Otherwise it is $\lambda_{\max}$ if $r=0$, or $\lambda_{\max}\min\{1,(e-e_w)/\max(1,\lfloor rT\rfloor)\}$. This is a configured epoch schedule, not a feedback-adaptive optimizer. Dataset and method overrides must be interpreted per run.}}

\newcommand{\pub}[1]{\textit{#1}}   
\newcommand{\repro}[1]{#1}           
\newcommand{\bfr}[1]{\textbf{#1}}   
\newcommand{\best}[1]{\underline{\textbf{#1}}} 
\section{Experimental Results}
\label{sec:experiments}

\subsection{Experimental Setup}

\begin{table}[!tbp]
	\centering
	\caption{Training hyperparameters per dataset.
		Method-level overrides (below the rule) apply on top of dataset
		defaults.
		\textit{wd} = weight decay, \textit{mom} = momentum,
		\textit{bs} = batch size.}
	\label{tab:hyperparams}
	\resizebox{\columnwidth}{!}{
		\begin{tabular}{llcccccr}
			\toprule
			Dataset & Arch & Opt & lr & wd & mom & bs & Epochs \\
			\midrule
			CelebA (R18)  & R18 & SGD  & 1e-4 & 1e-4 & 0.9 & 128 & 50  \\
			CelebA (R50)  & R50 & SGD  & 1e-4 & 1e-4 & 0.9 & 128 & 50  \\
			CelebA-MB     & R18 & SGD  & 1e-3 & 1e-4 & 0.8 &  64 & 50  \\
			UrbanCars     & R50 & SGD  & 1e-3 & 1e-3 & 0.1 & 128 & 60  \\
			Waterbirds    & R50 & SGD  & 1e-3 & 1e-4 & 0.9 & 128 & 300 \\
			\midrule
			\multicolumn{8}{l}{\textit{Method overrides (applied on top of dataset defaults):}} \\
			LfF / DebiAN   & ---  & Adam & 1e-4 & 0.0  & ---  & 256 & same \\
			JTT (CelebA)   & ---  & SGD  & 1e-5 & 0.1  & 0.9  & 128 & +1 stage \\
			JTT (UrbanCars)& ---  & SGD  & 1e-5 & 0.1  & 0.9  & 128 & +3 stage \\
			JTT (Waterbirds)& --- & SGD  & 1e-5 & 0.1  & 0.9  & 128 & +60 stage \\
			DFR            & ---  & ---  & same & same & ---  & same & +5 L-BFGS \\
			\bottomrule
		\end{tabular}
	}
\end{table}

\colortext{blue}{The small-scale audit covers CelebA-Std~\cite{liu2015celeba,sagawa2020groupdro} (Blond Hair/Male, ResNet-18 and ResNet-50~\cite{he2016resnet}), CelebA-MB (Smiling/Male and Young, eight joint evaluation groups), UrbanCars~\cite{li2023whacamole} (urban/country with background and co-occurring-object attributes), and Waterbirds~\cite{sagawa2020groupdro} (bird type/background). A separate ImageNet-1K~\cite{russakovsky2015imagenet} experiment introduces a synthetic watermark during training (Section~\ref{subsec:imagenet_exp}). We compare ERM, LfF, JTT, DFR, DebiAN, and GroupDRO, with and without BFR. Reported benchmark values summarize three seeds. CelebA-Std uses validation WGA selection, whereas CelebA-MB and UrbanCars use validation average accuracy for the recorded runs. This difference can distort WGA comparisons and does not arise from a principled lack of group labels: BFR requires these annotations. We preserve existing results rather than retrospectively selecting unrecorded checkpoints. Unified reimplementations differ materially from some published baselines, particularly LfF on CelebA-R50; gains describe these implementations, not superiority over the original methods.}

\colortext{blue}{Table~\ref{tab:hyperparams} lists dataset defaults and method overrides. BFR uses $\lambda=20$ by default and $\lambda=1$ for JTT, based on the reported sensitivity study rather than a general optimality rule. The centroid buffer uses $\alpha=0.9$; the default absolute warmup is five epochs, with fractional warmup/ramp/stop settings available in the implementation. Probes are scheduled every five epochs. Saved per-run schedules and checkpoint selection are necessary to reproduce a particular result; a rare-group or EMA sensitivity study is not included.}

\subsection{Backbone Geometry Audit}
\label{subsec:ibmi_audit}

\begin{table*}[!t]
	\centering
	\caption{\colortext{blue}{IBMI of trained backbones (reported 3-seed means and standard deviations; UrbanCars reports mean background/object values). IBMI is a class-confounded centroid-alignment summary, not actual feature reliance. DFR freezes its own first-stage backbone, which need not equal the separately trained ERM run. $^*$Low-IBMI cases discussed in the text. N/A marks an unavailable configuration; GroupDRO uses ResNet-50 for the reported CelebA-MB configuration.}}
	\label{tab:ibmi}
	\small
	\setlength{\tabcolsep}{4pt}
	\begin{tabular}{lccccc}
		\toprule
		Method
		& CelebA R18
		& CelebA R50
		& CelebA-MB R18
		& UrbanCars$^\dagger$ R50
		& Waterbirds R50 \\
		\midrule
		ERM
		& $0.404{\pm}0.008$
		& $0.559{\pm}0.004$
		& $0.274{\pm}0.010$
		& $0.217$ / $0.489$
		& $0.357{\pm}0.029$ \\
		LfF
		& $0.681{\pm}0.055$
		& $0.881{\pm}0.003$
		& $0.940{\pm}0.017$
		& $0.108$ / $0.421$
		& $0.650{\pm}0.059$ \\
		JTT
		& $0.609{\pm}0.010$
		& $0.647{\pm}0.016$
		& $0.013{\pm}0.014^*$
		& $0.158$ / $0.075$
		& $0.060{\pm}0.012^*$ \\
		DFR
		& $0.502{\pm}0.019$
		& $0.559{\pm}0.004$
		& $0.468{\pm}0.017$
		& $0.212$ / $0.474$
		& $0.331{\pm}0.004$ \\
		DebiAN
		& $0.513{\pm}0.033$
		& $0.896{\pm}0.008$
		& $0.951{\pm}0.014$
		& $0.089$ / $0.312$
		& $0.597{\pm}0.034$ \\
		GroupDRO
		& $0.466{\pm}0.023$
		& $0.650{\pm}0.004$
		& N/A$^{\ddagger}$
		& $0.044{\pm}0.014$ / $0.372{\pm}0.018$
		& $0.329{\pm}0.003$ \\
		\bottomrule
	\end{tabular}
	
	\vspace{0.4em}
	\footnotesize
	$^\dagger$UrbanCars reports IBMI for two spurious attributes
	(background-correlation / co-occurring-object), 3-seed mean.
	IBMI is evaluated on converged backbones using the BiasFlow probe.
\end{table*}

\colortext{blue}{IBMI describes the alignment of class centroids with the pooled attribute-centroid axis. Low values indicate weak alignment in this geometry; high values indicate co-alignment, which can arise from class composition alone. Neither establishes a clean or contaminated backbone in a causal sense.}

Table~\ref{tab:ibmi} reports IBMI for all baseline methods across all datasets and architectures. No output-level method systematically reduces backbone IBMI below ERM, and several make it substantially worse.

\colortext{blue}{\textbf{DFR/ERM provenance.} The earlier assertion that the DFR row equals ERM by construction was incorrect for this audit. The saved baseline sweep invokes DFR with its own first-stage training rather than an ERM checkpoint, and the trainer freezes that first-stage model only during head retraining. Separate training, checkpoint selection, and probe evaluation can therefore yield different IBMI. We retain the recorded values ($0.404$/$0.502$ on CelebA-R18, $0.274$/$0.468$ on CelebA-MB, and $0.357$/$0.331$ on Waterbirds for ERM/DFR), remove claims of shared-checkpoint identity, and do not interpret these differences as an effect of head retraining. The run configuration explains a source of mismatch, but a checkpoint-by-checkpoint identity/probe audit is still needed to certify numerical provenance across every table cell.}

\colortext{blue}{LfF and DebiAN have larger recorded IBMI than ERM on CelebA and Waterbirds: for example, the CelebA-R50 values are $0.881$ and $0.896$ versus $0.559$. This reports an alignment difference; without controlled axes or interventions it does not identify a mechanism such as increased encoding or prove that the original head relies more on the attribute. GroupDRO's outcome robustness can likewise coexist with pooled alignment.}

\colortext{blue}{JTT has low IBMI on Waterbirds ($0.060$ versus ERM $0.357$) and CelebA-MB ($0.013$ versus $0.274$). These are dataset-specific alignment outcomes, not anomalies to discard or evidence that JTT universally removes attribute information. A causal explanation would require controlled measurement beyond the present audit.}

\colortext{blue}{The audit shows that output performance and the reported centroid geometry do not move uniformly together. The independent frozen-backbone analysis is needed to assess whether these differences have a practical consequence under new head training.}

\subsection{AblGap Analysis}
\label{subsec:ablgap_results}

\begin{figure}[!t]
	\centering
	\includegraphics[width=\columnwidth]{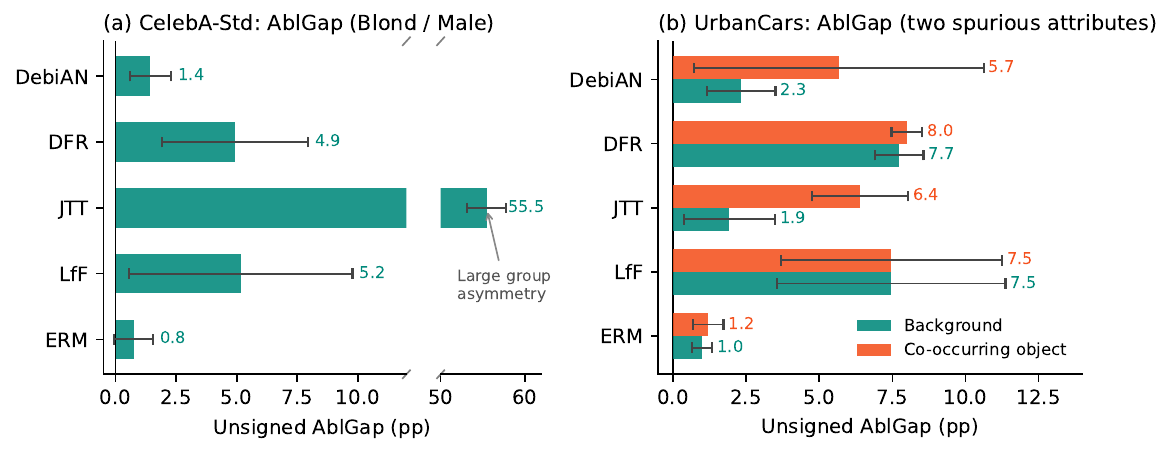}
	\caption{\colortext{blue}{Unsigned AblGap on CelebA-Std and UrbanCars, using the original plotted seed values. AblGap is $|\Delta\mathrm{Acc}_R-\Delta\mathrm{Acc}_U|$, reported in pp. Larger values indicate more asymmetric projection sensitivity across groups. The bars do not show the sign of either accuracy change, and cannot distinguish exploitation from interference.}}\label{fig:ablgap}
\end{figure}

\colortext{blue}{Figure~\ref{fig:ablgap} reports unsigned AblGap on converged baseline backbones for CelebA-Std and UrbanCars. The original plotting script reads the recorded AblGap fields, not signed $\Delta\mathrm{Acc}_R$. We retain those values and correct the axis description. AblGap measures the absolute difference in projection-induced accuracy changes between attribute groups.}

\colortext{blue}{JTT's CelebA-Std AblGap is $55.5$\,pp, a large asymmetry in group-specific projection sensitivity. All displayed bars are nonnegative by definition, so their signs cannot establish exploitation. Without the signed group drops, neither an exploitation nor an interference regime can be certified from this figure. Class-composition confounding further limits causal interpretation. The experiment documents projection sensitivity rather than validated intervention selection.}

\subsection{Comparison with Debiasing Baselines}
\label{subsec:bfr_results}

We now evaluate whether backbone-level geometric regularization translates to improved fairness across all datasets: CelebA-Std (Blond Hair target, Male spurious), CelebA-MB (Smiling target, Male + Young spurious), UrbanCars (scene classification, background + object spurious), and Waterbirds (bird type target, background spurious). We report Worst-Group Accuracy (WGA), the accuracy of the worst-performing subgroup defined by the cross-product of class label and spurious attribute. Higher WGA indicates fairer performance.

Tables~\ref{tab:bfr_celeba_r18}, \ref{tab:bfr_celeba_r50}, and~\ref{tab:bfr_scene} report WGA for all baselines with and without BFR-inv across all datasets and architectures. BFR improves or preserves WGA across all configurations, with the smallest gain observed for JTT on Waterbirds and average accuracy preserved within $0$--$2$\,pp throughout.

\colortext{blue}{BFR's reported gains vary by dataset and base method. DFR changes only its head during the second stage, whereas LfF, JTT, DebiAN, and GroupDRO can change their backbones throughout training. Their WGA objectives do not explicitly enforce BFR's centroid constraint. The observed gain ordering should not be read as a demonstrated correlation between IBMI and treatment benefit or as evidence that reweighting leaves representations unchanged.}

The relationship also explains why BFR is most valuable as a \emph{compositional} module rather than a standalone method. On CelebA-MB, where the multi-bias structure makes output-level corrections less effective, gains are largest precisely because no baseline method can address both spurious axes simultaneously through head compensation alone.

\noindent\textbf{Supervision.}
BFR requires spurious attribute labels and belongs to the same supervised category as GroupDRO~\cite{sagawa2020groupdro}. Gains over unsupervised methods (LfF, JTT, DebiAN) reflect both the geometric intervention and the additional supervision signal; absolute comparisons should be interpreted accordingly. A single hyperparameter rule $\lambda{=}20$ governs all configurations, reduced to $\lambda{=}1$ for JTT to prevent competing gradient signals with its $50{\times}$ sample upweighting; see the supplementary material for the $\lambda$ sensitivity sweep.
\begin{table*}[!t]
	\centering
	\caption{WGA (\%) on CelebA-Std (ResNet-18) and CelebA-MB (ResNet-18).
		3 seeds, mean$\pm$std.
			\textbf{Bold}: higher WGA between Reproduced and +BFR for each method.
		\pub{Italics}: published number for reference.
		$^\dagger$Supervised method.
		$^\ddagger$Published DFR (92.9) uses ResNet-50 + sklearn solver;
		our R18 reproduction uses L-BFGS head, explaining the gap.}
	\label{tab:bfr_celeba_r18}
	\small
	\setlength{\tabcolsep}{5pt}
	\begin{tabular}{ll ccc cc}
		\toprule
		& & \multicolumn{3}{c}{\textbf{CelebA-Std (ResNet-18)}}
		& \multicolumn{2}{c}{\textbf{CelebA-MB (ResNet-18)}} \\
		\cmidrule(lr){3-5}\cmidrule(lr){6-7}
		Method & Paradigm
		& Published WGA
		& Reproduced WGA
		& +BFR WGA
		& Reproduced WGA
		& +BFR WGA \\
		\midrule
		ERM
		& Baseline
		& \pub{37.9}
		& \repro{43.9$\pm$1.2}
			& \bfr{\textbf{54.6$\pm$2.6}}
		& \repro{40.3$\pm$1.6}
		& \bfr{\textbf{53.9$\pm$2.7}} \\
		
		LfF
		& Loss-reweight
		& \pub{43.3}
		& \repro{58.7$\pm$2.1}
			& \bfr{\textbf{61.7$\pm$2.3}}
		& \repro{49.1$\pm$3.3}
		& \bfr{\textbf{67.7$\pm$2.3}} \\
		
		JTT
		& Loss-reweight
		& \pub{81.1}
		& \repro{81.1$\pm$0.4}
			& \bfr{\textbf{82.2$\pm$0.5}}
		& \repro{62.7$\pm$3.8}
		& \bfr{\textbf{69.3$\pm$2.6}} \\
		
		DFR$^{\dagger\ddagger}$
		& Head-retrain
		& \pub{92.9}
		& \repro{80.6$\pm$2.5}
			& \bfr{\textbf{81.9$\pm$1.0}}
		& \repro{73.9$\pm$0.8}
		& \bfr{\textbf{75.7$\pm$1.0}} \\
		
		DebiAN
		& Loss-reweight
		& \pub{52.9}
		& \repro{64.4$\pm$3.2}
			& \bfr{\textbf{65.4$\pm$5.3}}
		& \repro{46.9$\pm$4.2}
		& \bfr{\textbf{58.4$\pm$2.4}} \\
		
		\midrule
		GroupDRO$^\dagger$
		& Supervised
		& \pub{88.9}
		& \repro{86.3$\pm$0.5}
		& \bfr{\textbf{88.0$\pm$1.5}}
		& \repro{55.7$\pm$0.5}
		& \bfr{\textbf{68.3$\pm$4.4}} \\
		
		\bottomrule
	\end{tabular}
	
	\vspace{0.3em}
	\footnotesize
	CelebA-Std reproduced numbers use WGA-based model selection
	(best validation worst-group accuracy), matching the standard
	benchmark protocol.
	CelebA-MB uses accuracy-based model selection.
	For CelebA-MB, all methods are compared within the same controlled
	implementation.
\end{table*}

\begin{figure}[!t]
\centering
\includegraphics[width=\columnwidth]{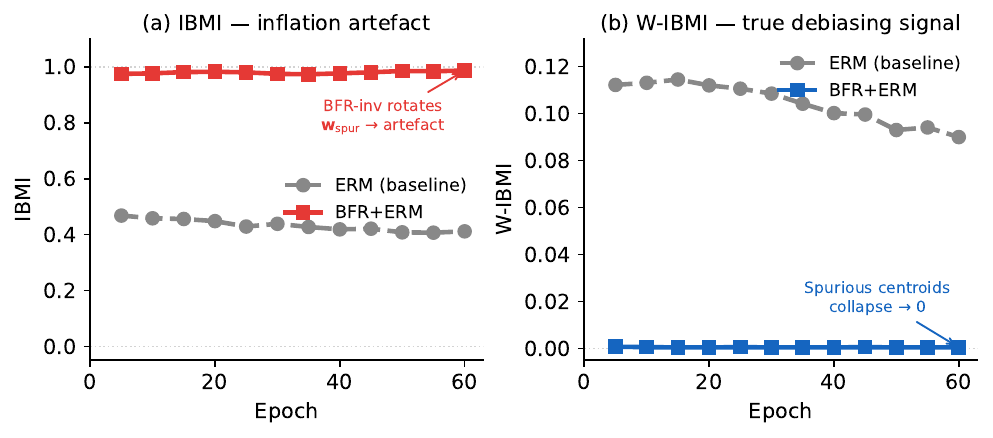}
\caption{\colortext{blue}{Recorded IBMI and W-IBMI trajectories for ERM and BFR+ERM on CelebA-Std. Pooled centroid alignment can increase as the attribute axis becomes dominated by class composition. W-IBMI tracks the centroid penalty being optimized, rather than independent concept erasure.}}
\label{fig:ibmi_trajectory}
\end{figure}

\noindent\textbf{W-IBMI tracks BFR's intended geometric effect.}
\colortext{blue}{W-IBMI is an objective-aligned verification metric: its reduction shows that the targeted first-moment discrepancy decreased under the chosen feature convention. Its formula matches BFR-inv apart from estimation details, so it is not independent evidence of attribute removal. Figure~\ref{fig:ibmi_trajectory} illustrates the pooled-axis limitation: IBMI can rise toward one while within-class centroid distance falls. Both summaries use model-dependent features; feature scale and class composition must be considered. Table~\ref{tab:bfr_wibmi} preserves the original measurements, including the BFR+GroupDRO R50 exception.}

\begin{table}[!tbp]
\centering
\caption{\colortext{blue}{Reported W-IBMI before and after BFR-inv (3-seed means). This scale-dependent first-moment statistic verifies an objective-aligned effect within each feature convention; it cannot rank architectures or certify erasure. $^*$BFR+GroupDRO R50 does not reduce W-IBMI; JTT+BFR Waterbirds retains a larger residual than most other configurations.}}
\label{tab:bfr_wibmi}
\small
\setlength{\tabcolsep}{4pt}
\begin{tabular}{l cc cc cc}
\toprule
 & \multicolumn{2}{c}{CelebA R18}
 & \multicolumn{2}{c}{CelebA R50}
 & \multicolumn{2}{c}{Waterbirds R50} \\
\cmidrule(lr){2-3}\cmidrule(lr){4-5}\cmidrule(lr){6-7}
Method & Base & +BFR & Base & +BFR & Base & +BFR \\
\midrule
ERM
  & $0.090$ & $0.001$
  & $0.018$ & $0.002$
  & $0.027$ & $0.001$ \\
LfF
  & $0.298$ & $0.000$
  & $0.082$ & $0.000$
  & $0.158$ & $0.000$ \\
JTT
  & $0.037$ & $0.001$
  & $0.008$ & $0.001$
  & $0.013$ & $0.004^{\ddagger}$ \\
DFR
  & $0.063$ & $0.001$
  & $0.018$ & $0.002$
  & $0.026$ & $0.001$ \\
DebiAN
  & $0.154$ & $0.001$
  & $0.060$ & $0.001$
  & $0.116$ & $0.001$ \\
GroupDRO
  & $0.074$ & $0.000$
  & $0.019$ & $0.022^*$
  & $0.024$ & $0.001$ \\
\bottomrule
\end{tabular}

\vspace{0.3em}
\footnotesize
$^\ddagger$BFR+JTT Waterbirds W-IBMI $= 0.004$ (vs.\ $\approx\!0.001$ for others),
consistent with partial centroid collapse under JTT's conflicting
$50{\times}$ upweighting schedule.
\end{table}

\begin{figure*}[t]
\centering
\includegraphics[width=\textwidth]{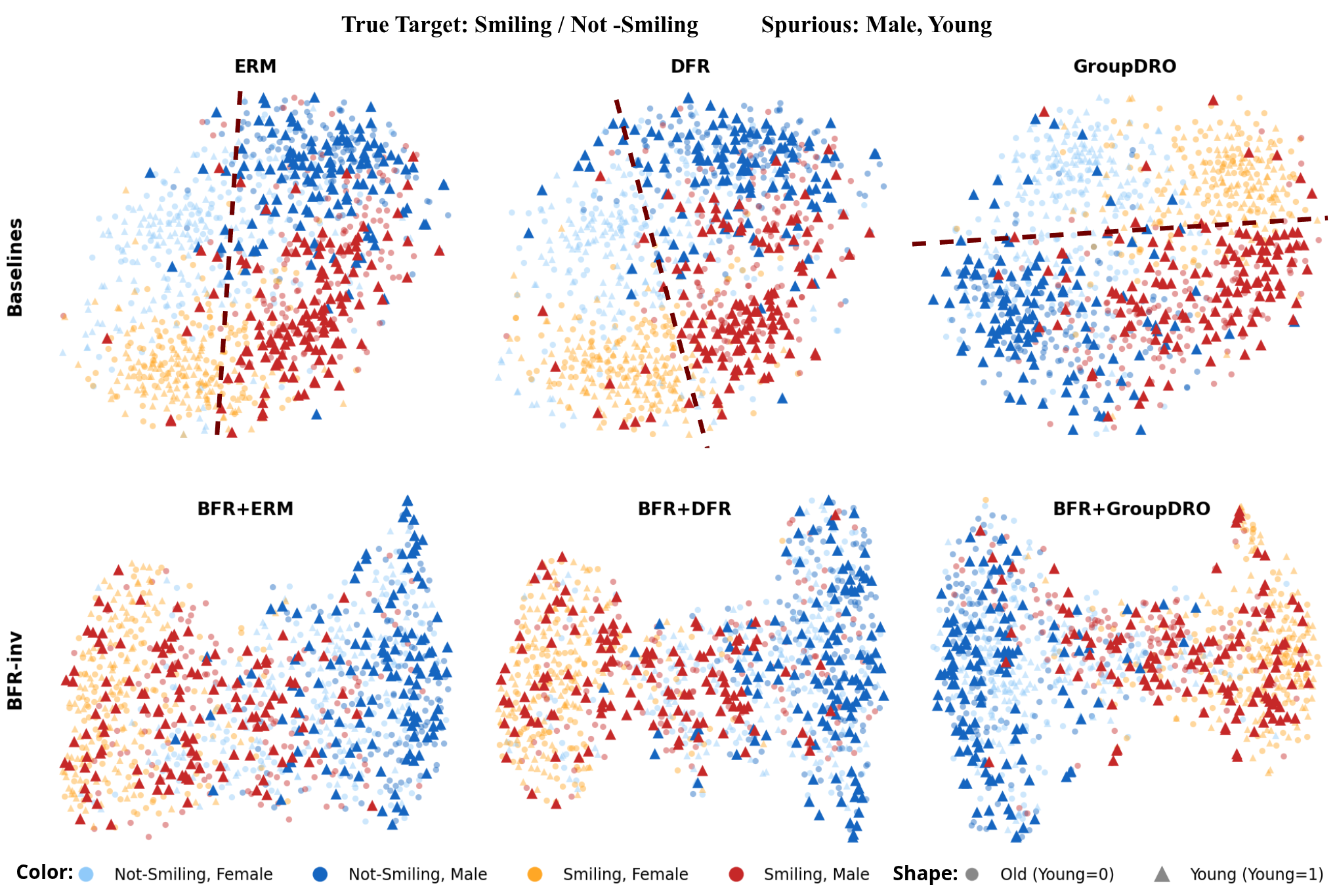}
\caption{\colortext{blue}{t-SNE of CelebA-MB ResNet-18 features (Smiling target; Male and Young annotated attributes). Color encodes target and Male; shape encodes Young. Baseline panels show subgroup separation; BFR panels illustrate more within-class mixing. BFR averages separate class-conditional penalties for the two supplied attributes. The visualization does not establish joint distribution alignment or lack of attribute recoverability.}}
\label{fig:tsne_mb}
\end{figure*}

\noindent\textbf{ResNet-50 results (Table~\ref{tab:bfr_celeba_r50}).}
\colortext{blue}{The CelebA-Std ResNet-50 configuration with ImageNet pretraining also has positive reported WGA gains, including LfF ($+6.7$\,pp) and DFR ($+4.1$\,pp). GroupDRO reaches $88.5\%$, below the separately published HierDRO value of $90.4\%$. This is an architecture-specific replication within the reported pipeline, not evidence of unrestricted generalization.}

\begin{table}[!tbp]
\centering
\caption{WGA (\%) on CelebA-Std (ResNet-50,
  ImageNet pretrained). 3 seeds, mean$\pm$std, WGA-based model selection.
	  \textbf{Bold}: higher WGA between Reproduced and +BFR for each method.
  \pub{Italics}: published number for reference.
  $^\dagger$Supervised method.}
\label{tab:bfr_celeba_r50}

\setlength{\tabcolsep}{5pt}

\begin{tabular}{ll ccc}
\toprule
& & \multicolumn{3}{c}{\textbf{WGA (\%)}} \\
\cmidrule(lr){3-5}
Method & Paradigm
  & Published
  & Reproduced
  & +BFR \\
\midrule
ERM
  & Baseline
  & \pub{47.7}
  & \repro{$49.1{\pm}1.6$}
	  & \bfr{$\mathbf{50.6{\pm}0.5}$} \\

LfF
  & Loss-reweight
  & \pub{77.2}
  & \repro{$55.9{\pm}5.9$}
	  & \bfr{$\mathbf{62.6{\pm}8.1}$} \\

JTT
  & Loss-reweight
  & \pub{81.5}
  & \repro{$80.2{\pm}0.2$}
  & \bfr{$\mathbf{82.1{\pm}0.2}$} \\

DFR$^\dagger$
  & Head-retrain
  & \pub{80.7}
  & \repro{$81.3{\pm}1.8$}
	  & \bfr{$\mathbf{85.4{\pm}2.2}$} \\

DebiAN
  & Loss-reweight
  & \pub{---}
  & \repro{$62.6{\pm}3.3$}
	  & \bfr{$\mathbf{64.1{\pm}3.0}$} \\

\midrule
GroupDRO$^\dagger$
  & Supervised
  & \pub{89.3}
  & \repro{$88.0{\pm}0.7$}
  & \bfr{$\mathbf{88.5{\pm}0.5}$} \\

\midrule
HierDRO$^\dagger$
  & Supervised ref.
  & \pub{90.4}
  & ---
  & --- \\

\bottomrule
\end{tabular}

\vspace{0.3em}
\footnotesize
All results use WGA-based model selection (best validation worst-group accuracy).
\colortext{blue}{DFR freezes its own ERM-trained first-stage backbone for L-BFGS head retraining. The audit does not establish identity with the separate ERM checkpoint.}
\end{table}

\begin{table*}[!tbp]
\centering
\caption{WGA (\%) and ID Accuracy (\%) on scene/object datasets
  (3 seeds, mean$\pm$std).
	  \textbf{Bold}: higher value between base and +BFR for each reported metric within each method.
  $^\dagger$Supervised method.}
\label{tab:bfr_scene}
\small
\setlength{\tabcolsep}{4pt}
\begin{tabular}{ll cccc}
\toprule
& & \multicolumn{2}{c}{\textbf{UrbanCars (R50)}}
  & \multicolumn{2}{c}{\textbf{Waterbirds (R50)}} \\
\cmidrule(lr){3-4}\cmidrule(lr){5-6}
Method & Paradigm
  & ID Acc (base / +BFR)
  & WGA (base / +BFR)
  & Avg Acc (base / +BFR)
  & WGA (base / +BFR) \\
\midrule
ERM
  & Baseline
  & $\mathbf{97.5{\pm}0.3}$ / $96.4{\pm}0.6$
  & $56.7{\pm}0.5$ / \bfr{$\mathbf{82.7{\pm}0.5}$}
  & $91.2$ / \bfr{$\mathbf{94.0}$}
  & $79.0{\pm}0.1$ / \bfr{$\mathbf{83.2{\pm}1.4}$} \\

LfF
  & Loss-reweight
  & $97.2{\pm}1.7$ / \bfr{$\mathbf{97.4{\pm}0.9}$}
  & $67.7{\pm}3.6$ / \bfr{$\mathbf{85.6{\pm}2.8}$}
  & $91.9$ / \bfr{$\mathbf{92.7}$}
  & $79.4{\pm}1.5$ / \bfr{$\mathbf{86.5{\pm}1.6}$} \\

JTT$^\ast$
  & Loss-reweight
  & $91.6{\pm}1.4$ / \bfr{$\mathbf{91.9{\pm}1.7}$}
	  & $56.7{\pm}2.5$ / \bfr{$\mathbf{60.1{\pm}2.3}$}
  & $88.1$ / \bfr{$\mathbf{88.3}$}
  & $80.8{\pm}3.8$ / \bfr{$\mathbf{80.9{\pm}3.8}$} \\

DFR$^\dagger$
  & Head-retrain
  & $86.8{\pm}0.6$ / \bfr{$\mathbf{88.4{\pm}0.1}$}
  & $78.1{\pm}0.2$ / \bfr{$\mathbf{83.9{\pm}1.6}$}
  & $\mathbf{93.3}$ / $93.0$
	  & $91.3{\pm}0.5$ / \bfr{$\mathbf{91.4{\pm}0.5}$} \\

DebiAN
  & Loss-reweight
  & $\mathbf{98.1{\pm}1.1}$ / $97.5{\pm}0.3$
  & $63.7{\pm}2.6$ / \bfr{$\mathbf{79.6{\pm}2.6}$}
  & $91.0$ / \bfr{$\mathbf{93.8}$}
  & $80.1{\pm}2.4$ / \bfr{$\mathbf{87.9{\pm}1.3}$} \\

\midrule
GroupDRO$^\dagger$
  & Supervised
  & $74.1{\pm}1.2$ / \bfr{$\mathbf{87.5{\pm}0.2}$}
  & $74.1{\pm}1.2$ / \bfr{$\mathbf{87.5{\pm}0.2}$}
  & $94.0$ / \bfr{$\mathbf{94.2}$}
	  & $88.8{\pm}1.1$ / \bfr{$\mathbf{89.9{\pm}0.5}$} \\

\bottomrule
\end{tabular}

\vspace{0.3em}
\footnotesize
Waterbirds uses val-WGA model selection (standard benchmark protocol).
$^\ast$BFR+JTT uses $\lambda{=}1$ to avoid gradient conflict with JTT's $50{\times}$ upweighting on the 300-epoch schedule.
\end{table*}

\noindent\textbf{Scene and object datasets (Table~\ref{tab:bfr_scene}).}
\colortext{blue}{UrbanCars has the largest reported WGA gain ($+26.0$\,pp for BFR+ERM). On Waterbirds the mean gains are positive, but BFR+JTT is effectively neutral ($80.9\%$ versus $80.8\%$); the appendix records sensitivity to stronger regularization. The preserved table results characterize the tested pipelines, not a verified single-direction mechanism or universal accuracy preservation.}

\colortext{blue}{Table~\ref{tab:bfr_wibmi} shows substantial reductions in the targeted centroid statistic in most settings, with the explicit BFR+GroupDRO R50 exception and a Waterbirds JTT residual of $0.004$. These are first-moment changes under a fixed reporting convention. Figure~\ref{fig:tsne_mb} is a qualitative visualization of attribute-wise supervised alignment, not proof that all four joint subgroups have identical distributions. The frozen-head and attribute-probe experiments assess practical consequences independently of this loss-aligned statistic.}

\subsection{Comparison with Supervised Methods}
\label{subsec:gdro_results}

Since BFR requires spurious attribute labels at training time, it belongs to the same supervised category as GroupDRO~\cite{sagawa2020groupdro}. Table~\ref{tab:gdro} reports results when BFR is applied on top of GroupDRO directly.

\colortext{blue}{The BFR+GroupDRO comparison keeps training-time attribute supervision in both methods and is therefore more informative about the added penalty than comparisons with unannotated baselines. Recorded gains are largest on CelebA-MB ($+12.5$\,pp) and UrbanCars ($+13.3$\,pp). GroupDRO trains the whole model, while BFR imposes an additional centroid constraint; the gains do not establish separate head-only and backbone-only failure modes.}

\begin{table}[!tbp]
	\centering
	\caption{GroupDRO vs. BFR+GroupDRO (3 seeds, mean$\pm$std). 
		WGA (\%). \colortext{blue}{BFR adds gains to the evaluated supervised GroupDRO baseline.}}
	\label{tab:gdro}
	\setlength{\tabcolsep}{4pt} 
	\begin{tabular}{lcccc}
		\toprule
		Method & CelebA-Std & CelebA-Std & CelebA-MB & UrbanCars \\
		& (R18)      & (R50)      & (R18)     & (R50)     \\
		\midrule
		GroupDRO
		& $86.3{\pm}0.5$
		& $88.0{\pm}0.7$
		& $55.7{\pm}0.5$
		& $74.1{\pm}1.2$ \\
		BFR + GroupDRO
		& $\mathbf{88.0{\pm}1.5}$
		& $\mathbf{88.5{\pm}0.5}$
		& $\mathbf{68.3{\pm}4.4}$
		& $\mathbf{87.5{\pm}0.2}$ \\
		\midrule
		$\Delta$ WGA (pp)
		& $+1.7$
		& $+0.5$
		& $+12.5$
		& $+13.3$ \\
		\bottomrule
	\end{tabular}

	\footnotesize
	CelebA-Std uses WGA-based model selection; CelebA-MB and UrbanCars use accuracy-based selection. \\
	Waterbirds GroupDRO ($88.8{\pm}1.1$ base, $89.9{\pm}0.5$ +BFR, $\Delta{+}1.1$\,pp) is included in Table~\ref{tab:bfr_scene}.
\end{table}
\colortext{blue}{\subsection{Controlled Synthetic-Watermark ImageNet Study}}
\label{subsec:imagenet_exp}

\colortext{blue}{CelebA uses demographic attributes, while UrbanCars and Waterbirds use contextual attributes. We separately test a controlled synthetic watermark on ImageNet-1K to assess a known training-time nuisance under matched ERM and BFR training. This is a supervised synthetic-spurious-feature experiment.}

\noindent\textbf{Setup.}
\colortext{blue}{The reported experiment injects a watermark into ImageNet-1K training images with probability $p=0.5$ and trains matched ResNet-50 ERM and BFR+ERM models from scratch for 90 epochs. Watermark-related evaluation is motivated by ImageNet-W~\cite{li2023whacamole} and SEBRA~\cite{kappiyath2025sebra}. Because training includes an explicitly injected, annotated watermark, this is not the canonical evaluation of naturally trained ImageNet models on ImageNet-W. We retain the reported IN-W label for the evaluation split, but make no directly comparable canonical-benchmark or state-of-the-art claim. Table~\ref{tab:imagenet_ood} also reports clean validation and four additional shifts.}

\noindent\textbf{Results.}
\colortext{blue}{Under this controlled training protocol, IN-W accuracy increases from $52.6\%$ to $75.6\%$ ($+23.0$\,pp), while clean validation accuracy remains $76.0\%$. The four other reported shifts change by less than $2$\,pp. W-IBMI decreases from $0.113$ to $<0.001$ and unsigned AblGap from $0.161$ to $0.001$; these are, respectively, objective-aligned centroid distance and group-asymmetric projection sensitivity, not two independent demonstrations of watermark erasure. The geometry rows contain point summaries without reported per-seed uncertainty, unlike the accuracy rows. The independent accuracy comparison supports utility under this synthetic intervention, but not general improvement on unrelated OOD shifts.}

\begin{table}[!tbp]
\centering
\caption{\colortext{blue}{Controlled synthetic-watermark training on ImageNet-1K, ResNet-50. Accuracy rows retain 3-seed means and standard deviations; IN-W denotes the reported watermark-shift evaluation after injected-watermark training, not a canonical natural-training comparison. Geometry rows retain point summaries: W-IBMI is objective aligned and scale dependent; AblGap is unsigned. Accuracy changes are in pp.}}
\label{tab:imagenet_ood}
\setlength{\tabcolsep}{6pt}
\begin{tabular}{lccc}
\toprule
& ERM & BFR & $\Delta$ \\
\midrule
\multicolumn{4}{l}{\textit{OOD accuracy} (\%, $\uparrow$)} \\
Val (ID)      & $76.0{\pm}0.1$ & $76.0{\pm}0.1$ & $\phantom{+}0.0$ \\
IN-W          & $52.6{\pm}0.6$ & $\mathbf{75.6{\pm}0.1}$ & $+23.0$ \\
IN-R          & $36.2{\pm}0.6$ & $36.5{\pm}0.2$ & $+0.3$ \\
IN-Stylized   & $52.1{\pm}1.0$ & $53.2{\pm}0.5$ & $+1.1$ \\
mixed\_same   & $87.2{\pm}0.3$ & $87.3{\pm}0.1$ & $+0.1$ \\
mixed\_rand   & $80.7{\pm}0.6$ & $81.3{\pm}0.5$ & $+0.6$ \\
\midrule
\multicolumn{4}{l}{\textit{BiasFlow geometry} ($\downarrow$)} \\
W-IBMI        & $0.113$        & $\mathbf{<0.001}$       & $-0.113$ \\
AblGap        & $0.161$        & $\mathbf{0.001}$        & $-0.160$ \\
\bottomrule
\end{tabular}

\end{table}

\noindent\textbf{Implication.}
\colortext{blue}{The synthetic study provides a matched large-scale example in which BFR improves watermark-shift accuracy while preserving the reported clean accuracy. Canonical ImageNet-W evaluation without synthetic watermark injection, and comparisons with equally supervised representation-level methods, remain separate experiments.}

\subsection{Backbone and Head Contribution Analysis}
\label{subsec:exp2_probe}

\colortext{blue}{We freeze the converged CelebA-Std ResNet-18 backbones and learn new linear heads. Balanced mode uses group-balanced validation examples with L-BFGS; biased mode uses the standard training distribution with SGD; attribute-probe mode trains an SGD linear classifier on validation features to predict Male and evaluates it on the test set. This is the saved probe implementation, rather than a separately fitted logistic-regression solver. Table~\ref{tab:exp2_probe} reports three-seed means, with variability reconstructed from saved per-seed outputs in Appendix~\ref{app:probe_details}. Balanced and biased modes differ in data and solver, so their difference is a protocol stress test, not a causal decomposition or an optimal-head bound.}

\colortext{blue}{\textbf{Balanced mode} yields approximately 79--84\% WGA across the evaluated backbones under the supplied balancing and solver. This shows useful target signal is available under this protocol; it does not establish the best possible head or representation invariance.}

\colortext{blue}{\textbf{Biased mode} supplies the principal independent test, since it does not reuse the centroid formula. GroupDRO's frozen backbone yields $40.7\%$ WGA with a fresh biased head despite $86.3\%$ with its original head. BFR+GroupDRO yields $64.1\%$ with the fresh biased head, a $23.4$\,pp gain over GroupDRO, while BFR+LfF gains $15.2$\,pp over LfF. The strongest increment is therefore GroupDRO, not LfF. Other increments include the smaller DFR change ($42.2\%$ to $44.4\%$). The results support reduced susceptibility to this particular head-training procedure. Changes in optimization, margins, and available task signal can also affect WGA, so the original-head drop is not pure head compensation or a causal attribution to Male.}

\colortext{blue}{\textbf{Attribute-probe mode} finds Male test accuracy of roughly 90--93\% for the base backbones, and lower accuracy after BFR. However, Male remains recoverable: BFR+ERM, BFR+JTT, BFR+DFR, and BFR+GroupDRO obtain $80.0\%$, $86.6\%$, $67.5\%$, and $72.2\%$. We report differences descriptively without a statistical-equivalence claim. Raw accuracy is affected by label imbalance, class correlation, and probe optimization; a class-balanced or class-conditional test would more directly separate residual encoding from composition. Attribute availability to a new probe does not show that the original classifier uses it.}

\begin{table}[!t]
\centering
\caption{\colortext{blue}{Frozen-backbone stress tests on CelebA-Std (R18), three-seed means. Biased WGA is test WGA of a fresh SGD task head fitted on biased training features. Male Probe is test accuracy of an SGD linear attribute classifier fitted on validation features. Each pair is base / +BFR; bold marks higher WGA or lower attribute-probe accuracy. Lower probe accuracy indicates reduced recoverability under this protocol, not certified erasure. Per-seed variability is in Appendix~\ref{app:probe_details}.}}
\label{tab:exp2_probe}
\footnotesize
\setlength{\tabcolsep}{3pt}
\begin{tabular}{llcc}
\toprule
\multirow{2}{*}{Method} & \multirow{2}{*}{Paradigm} & Biased WGA ($\uparrow$) & Male Probe ($\downarrow$) \\
                        &                           & (base / +BFR)           & (base / +BFR) \\
\midrule
ERM      & Baseline     & $40.0$ / \bfr{$\mathbf{44.6}$} & $93.2$ / \bfr{$\mathbf{80.0}$} \\
LfF      & Loss-rewt.   & $44.8$ / \bfr{$\mathbf{60.0}$} & $90.6$ / \bfr{$\mathbf{61.4}$} \\
JTT      & Loss-rewt.   & $36.5$ / \bfr{$\mathbf{42.2}$} & $93.4$ / \bfr{$\mathbf{86.6}$} \\
DFR      & Head-retrain & $42.2$ / \bfr{$\mathbf{44.4}$} & $91.9$ / \bfr{$\mathbf{67.5}$} \\
DebiAN   & Loss-rewt.   & $45.4$ / \bfr{$\mathbf{52.6}$} & $92.8$ / \bfr{$\mathbf{65.2}$} \\
GroupDRO & Supervised   & $40.7$ / \bfr{$\mathbf{64.1}$} & $92.5$ / \bfr{$\mathbf{72.2}$} \\
\bottomrule
\end{tabular}
\end{table}

\section{Discussion and Analysis}
\label{sec:discussion}

\noindent\textbf{WGA and backbone geometry are decoupled.}
\colortext{blue}{A predictor's WGA and the performance of a new head on its frozen backbone characterize different evaluation procedures. GroupDRO's $86.3\%$ original-head WGA and $40.7\%$ biased-head WGA illustrate this distinction; BFR+GroupDRO increases the latter to $64.1\%$. This supports reporting fresh-head stress tests alongside outcome metrics. Continued-training and Smiling-transfer results are mixed and do not certify robustness under arbitrary fine-tuning, tasks, or shifts.}

\colortext{blue}{\noindent\textbf{Alignment differences do not establish a mechanism.}}
\colortext{blue}{LfF and DebiAN have higher reported pooled IBMI than ERM in several configurations. This can reflect changes in class geometry, attribute variation, or the model-specific axis. The audit does not isolate those explanations or show that reweighting intrinsically worsens representations. Stronger representation-level baselines and fixed or conditional axes are needed to assess this question.}

\noindent\textbf{Backbone geometry scales to large benchmarks.}
\colortext{blue}{The controlled synthetic-watermark experiment supports the utility of the supervised penalty under matched large-scale training: watermark-shift accuracy improves by $23.0$\,pp, with clean accuracy unchanged at the reported precision. Its deliberately altered training protocol limits comparison with canonical ImageNet-W results. The small changes on four other shifts do not establish general OOD robustness.}

\noindent\textbf{Metric scope and limitations.}
\colortext{blue}{IBMI measures centroid alignment, W-IBMI measures a scale-dependent centroid penalty, and AblGap measures asymmetric sensitivity to projection. They do not measure total attribute information or certify fairness. Class-composition confounding applies to IBMI and projection sensitivity even before intervention. Low W-IBMI does not imply linear unrecoverability, as the Male probes demonstrate. Interference-like negative projection drops and diagnostic mode selection are not experimentally validated here. BFR requires all targeted attribute labels; attribute-wise penalties do not guarantee joint invariance. Detached EMA fallback can become stale and gives no gradient for absent groups when both centroids come from memory. Rare-group sampling, EMA sensitivity, balanced independent probes, and normalized geometry remain untested. Model-selection differences and baseline reproduction gaps restrict comparative claims. The strongest independent evidence is same-task biased-head resistance; cross-task utility and information preservation must be evaluated for each target task.}

\colortext{blue}{BiasFlow combines lightweight centroid diagnostics with projection sensitivity and fresh-head evaluation. The retained experiments show that WGA alone does not determine frozen-backbone behavior under biased head retraining. BFR, a supervised class-conditional centroid-alignment penalty related to existing representation-learning work, improves or preserves the reported mean WGA across the evaluated configurations and improves biased-head WGA most clearly for GroupDRO on CelebA-Std. W-IBMI verifies the optimized first-moment discrepancy, while probes show reduced but persistent attribute recoverability. The synthetic-watermark ImageNet study supports a matched nuisance intervention rather than canonical ImageNet-W superiority. Representation-level baseline comparisons, controlled metric validation, consistent checkpoint selection, rare-group analysis, and broader transfer evaluation remain necessary to substantiate stronger claims.}

\clearpage
\onecolumn
\appendices
\setcounter{section}{0}
\setcounter{figure}{0}
\setcounter{table}{0}
\setcounter{equation}{0}
\renewcommand{\theHsection}{appendix.\arabic{section}}
\renewcommand{\theHfigure}{appendix.\arabic{figure}}
\renewcommand{\theHtable}{appendix.\arabic{table}}
\renewcommand{\theHequation}{appendix.\arabic{equation}}
\renewcommand{\thesection}{\Alph{section}}
\renewcommand{\thesubsection}{\thesection.\arabic{subsection}}
\renewcommand{\thefigure}{S\arabic{figure}}
\renewcommand{\thetable}{S\arabic{table}}
\renewcommand{\theequation}{S\arabic{equation}}
\section*{\colortext{blue}{Appendix: Additional Methods and Stress Tests}}
\input{appendix/biasflow_appendix_content.tex}

\clearpage
{\small
\bibliographystyle{IEEEtran}
\bibliography{reference}
}

\end{document}

%% file: plots/framework.tex
\usetikzlibrary{shapes.geometric}
\definecolor{BFnavy}{HTML}{263D52}
\definecolor{BFblue}{HTML}{2278AA}
\definecolor{BForange}{HTML}{D66D2D}
\definecolor{BFteal}{HTML}{23887E}
\begin{tikzpicture}[x=1cm,y=1cm,every node/.style={font=\footnotesize,align=center},
 panel/.style={draw=BFnavy!25,rounded corners=3pt,fill=BFnavy!2},
 flow/.style={->,>=stealth,thick,draw=BFnavy!75},
 circ/.style={circle,fill=BFblue,draw=white,inner sep=1.5pt},
 tri/.style={regular polygon,regular polygon sides=3,fill=BForange,draw=white,inner sep=1.3pt},
 cent/.style={circle,draw=BFnavy,fill=white,inner sep=1.4pt,line width=.6pt},
 chip/.style={draw=BFnavy!45,fill=white,rounded corners=2pt,inner sep=4pt}]
\path[use as bounding box] (0,0) rectangle (18.1,7.2);
\node[anchor=west,text=BFnavy] at (0,6.96) {\colortext{blue}{\textbf{Training data} $(x,y,s_1,\ldots,s_K)$}};
\draw[flow] (4.7,6.96)--(5.45,6.96);
\node[chip,text=BFnavy] at (7.15,6.96) {\colortext{blue}{Backbone $f_\theta(x)=z$}};
\draw[flow] (8.84,6.96)--(9.55,6.96);
\node[chip,text=BFnavy] at (11.0,6.96) {\colortext{blue}{Task head $h(z)$}};
\draw[flow] (12.2,6.96)--(12.8,6.96);
\node[anchor=west,text=BFnavy] at (12.83,6.96) {\colortext{blue}{Prediction $\hat y$; base training loss}};
\foreach \x in {0,6.15,12.3} {\draw[panel] (\x,.62) rectangle ++(5.8,5.82);}
\node[anchor=west,font=\small\bfseries,text=BFnavy] at (.22,6.13) {\colortext{blue}{(a) Monitor feature geometry}};
\node[anchor=west,font=\small\bfseries,text=BFnavy] at (6.37,6.13) {\colortext{blue}{(b) Regularize within each class}};
\node[anchor=west,font=\small\bfseries,text=BFnavy] at (12.52,6.13) {\colortext{blue}{(c) Test the frozen backbone}};
\node[font=\scriptsize,text=BFnavy!85] at (2.9,5.74) {\colortext{blue}{Forward hooks capture backbone features}};
\begin{scope}[shift={(.5,3.55)},y=.83cm]
\draw[->,draw=BFnavy!40] (0,0)--(4.72,0) node[below] {$z_1$};
\draw[->,draw=BFnavy!40] (0,0)--(0,2.52) node[left] {$z_2$};
\draw[BFnavy!30,dashed] (1.38,.75) ellipse (1.05 and .6);
\draw[BFnavy!30,dashed] (3.35,1.92) ellipse (1.03 and .6);
\node[text=BFnavy] at (.65,.13) {\colortext{blue}{$y=0$}};
\node[text=BFnavy] at (4.35,2.37) {\colortext{blue}{$y=1$}};
\foreach \x/\y in {.75/.65,.88/.83,1.03/.6,2.72/1.8,2.9/1.99,3.04/1.76} {\node[circ] at (\x,\y) {};}
\foreach \x/\y in {1.75/.88,1.92/1.04,2.07/.8,3.72/2.03,3.89/2.2,4.04/1.96} {\node[tri] at (\x,\y) {};}
\draw[<->,>=stealth,thick,draw=BFnavy] (.9,.73)--(1.9,.91);
\node[cent] at (.9,.73) {}; \node[cent] at (1.9,.91) {};
\draw[<->,>=stealth,thick,draw=BFnavy] (2.89,1.85)--(3.89,2.05);
\node[cent] at (2.89,1.85) {}; \node[cent] at (3.89,2.05) {};
\node[fill=BFnavy!2,inner sep=1pt] at (1.5,1.27) {\colortext{blue}{$\Delta_0$}};
\node[fill=BFnavy!2,inner sep=1pt] at (3.47,1.52) {\colortext{blue}{$\Delta_1$}};
\end{scope}
\node[text width=5.24cm] at (2.9,2.66) {\colortext{blue}{\textbf{IBMI:} pooled class--attribute alignment\\\textbf{AblGap:} unsigned projection asymmetry}};
\node[text width=5.1cm,text=BFnavy!85,font=\scriptsize] at (2.9,1.7) {\colortext{blue}{Pooled geometry can reflect class composition.\\Projection sensitivity is not causal feature use.}};
\node[font=\scriptsize,text=BFnavy!85] at (9.05,5.74) {\colortext{blue}{BFR-inv targets conditional first moments}};
\begin{scope}[shift={(6.65,3.55)},y=.83cm]
\draw[->,draw=BFnavy!40] (0,0)--(4.72,0) node[below] {$z_1$};
\draw[->,draw=BFnavy!40] (0,0)--(0,2.52) node[left] {$z_2$};
\draw[BFnavy!30,dashed] (1.35,.75) ellipse (.77 and .6);
\draw[BFnavy!30,dashed] (3.36,1.92) ellipse (.77 and .6);
\node[text=BFnavy] at (.62,.13) {\colortext{blue}{$y=0$}};
\node[text=BFnavy] at (4.35,2.37) {\colortext{blue}{$y=1$}};
\foreach \x/\y in {1.08/.58,1.29/.93,1.52/.62,3.09/1.75,3.30/2.10,3.53/1.79} {\node[circ] at (\x,\y) {};}
\foreach \x/\y in {1.18/.93,1.41/.58,1.65/.93,3.19/2.10,3.42/1.75,3.66/2.10} {\node[tri] at (\x,\y) {};}
\node[cent] at (1.30,.74) {}; \node[cent] at (1.43,.82) {};
\node[cent] at (3.31,1.91) {}; \node[cent] at (3.44,1.99) {};

\end{scope}
\node[text width=5.36cm] at (9.05,2.57) {\colortext{blue}{$\displaystyle L_{\mathrm{inv}}=\frac{1}{C}\sum_c\frac{\|\mu_{c,1}-\mu_{c,0}\|^2}{d}$\\[4pt]$\displaystyle L=L_{\mathrm{base}}+\lambda\frac{1}{K}\sum_k L_{\mathrm{inv}}^{(k)}$}};
\node[text width=5.14cm,font=\scriptsize,text=BFnavy!85] at (9.05,1.40) {\colortext{blue}{Train $f_\theta$ using class and attribute labels.\\Match first moments, not full distributions.}};
\node[chip,text=BFnavy,text width=4.5cm] (frozen) at (15.2,5.4) {\colortext{blue}{\textbf{Freeze $f_\theta$}\quad fixed features $z$}};
\draw[flow] (15.2,5.03)--(15.2,4.69)--(13.88,4.69)--(13.88,4.38);
\draw[flow] (15.2,4.69)--(16.55,4.69)--(16.55,4.38);
\node[chip,text width=2.05cm,minimum height=1.17cm] (task) at (13.88,3.74) {\colortext{blue}{\textbf{Fresh task head}\\Train on biased\\training features}};
\node[chip,text width=2.05cm,minimum height=1.17cm] (probe) at (16.55,3.74) {\colortext{blue}{\textbf{Attribute probe}\\Train on validation\\features}};
\draw[flow] (task.south)--(13.88,2.80);
\draw[flow] (probe.south)--(16.55,2.80);
\node[text width=2.15cm,text=BFteal] at (13.88,2.47) {\colortext{blue}{Original-task\\test WGA}};
\node[text width=2.15cm,text=BFteal] at (16.55,2.47) {\colortext{blue}{Test attribute\\accuracy}};
\node[text width=5.05cm,font=\scriptsize,text=BFnavy!85] at (15.2,1.60) {\colortext{blue}{Test susceptibility to biased head training.\\Measure remaining attribute recoverability.}};
\node[fill=BFteal!10,rounded corners=2pt,text width=10.96cm,inner sep=4pt,text=BFnavy] at (5.98,.87) {\colortext{blue}{\textbf{W-IBMI:} objective-aligned verification; scale dependent}};
\node[fill=BFteal!10,rounded corners=2pt,text width=5.07cm,inner sep=4pt,text=BFnavy] at (15.2,.87) {\colortext{blue}{\textbf{Predictive evidence:} WGA and new heads}};
\node[circ] at (.23,.20) {}; \node[anchor=west] at (.38,.2) {\colortext{blue}{$s=0$}};
\node[tri] at (1.65,.20) {}; \node[anchor=west] at (1.8,.2) {\colortext{blue}{$s=1$}};
\node[cent] at (3.10,.2) {}; \node[anchor=west] at (3.25,.2) {\colortext{blue}{Group centroid; $\Delta_c=\mu_{c,1}-\mu_{c,0}$}};
\node[anchor=east,font=\scriptsize,text=BFnavy!75] at (18.1,.20) {\colortext{blue}{Schematic geometry; no measured points or distances}};
\end{tikzpicture}

%% file: appendix/biasflow_appendix_content.tex
\section{Additional BFR Variants}
\label{app:bfr_variants}

The main paper uses BFR-inv as the geometric regularization penalty, defined as
\begin{equation}
\label{eq:inv-loss-recap}
L_{\mathrm{inv}} \;=\; \frac{1}{C} \sum_{c=1}^{C} \frac{\|\mu(y{=}c, s{=}1) - \mu(y{=}c, s{=}0)\|^2}{d},
\end{equation}
which collapses within-class spurious centroid separation in the backbone feature space ($d$ is the feature dimension). We also developed two alternative variants, BFR-orth and BFR-spectral, that target different aspects of backbone spurious geometry. Their definitions are included here for completeness, and a head-to-head WGA comparison against BFR-inv follows in Section~\ref{app:mode_ablation}.

\subsection{Orthogonality Loss (BFR-orth)}
The \texttt{orth} mode is the most lightweight geometric intervention. It penalizes the alignment between the class-discriminative direction ($d_{\mathrm{cls}} = \mu(y=1) - \mu(y=0)$) and the spurious direction ($d_{\mathrm{spur}} = \mu(s=1) - \mu(s=0)$):
\begin{equation}
\label{eq:orth-loss}
L_{\mathrm{orth}} \;=\; \cos^2(d_{\mathrm{cls}}, d_{\mathrm{spur}}) \;=\; \left( \frac{\langle d_{\mathrm{cls}}, d_{\mathrm{spur}} \rangle}{\|d_{\mathrm{cls}}\| \cdot \|d_{\mathrm{spur}}\|} \right)^2\,.
\end{equation}
\colortext{blue}{\textbf{Mechanism and scope.} BFR-orth penalizes pooled centroid-direction alignment. Because the attribute direction can include class-composition effects, minimizing this quantity may also suppress useful target geometry. The reported ablation does not establish a regime-selection rule or a general preference for this mode in multi-attribute settings.}

\subsection{Spectral Variance Loss (BFR-spectral)}
Unlike \texttt{orth} (which only targets class-boundary alignment) and \texttt{inv} (which targets class-conditional centroids), the \texttt{spectral} mode penalizes the fraction of \emph{total} feature variance that aligns with the spurious direction:
\begin{equation}
\label{eq:spectral-loss}
L_{\mathrm{spectral}} \;=\; \frac{\operatorname{Var}(z \cdot w_{\mathrm{spur}})}{\operatorname{Var}_{\mathrm{total}}(z)}\,,
\end{equation}
where $\operatorname{Var}_{\mathrm{total}}(z)$ is the total per-sample variance after mean-centering. \\
\colortext{blue}{\textbf{Mechanism and scope.} BFR-spectral penalizes variance along the estimated attribute-associated direction. Its empirical failures in the mode ablation show that removing such variance can damage target prediction. It has no demonstrated interference-removal guarantee.}

\section{BFR Regularization Mode Ablation}
\label{app:mode_ablation}

\noindent\textbf{BFR regularization mode.}
Table~\ref{tab:mode_ablation} compares the three BFR regularization modes on CelebA-Std R18 (seed=1). \texttt{inv} strictly dominates \texttt{orth} and \texttt{spectral} on WGA across all methods. \texttt{orth} only requires the class-boundary direction to be orthogonal to the spurious axis, leaving within-class spurious information intact; a linear head trained on biased data can still find and exploit this information through other feature directions. \texttt{inv} directly collapses within-class spurious structure, preventing any head from exploiting it regardless of training signal. \texttt{spectral} achieves the strongest IBMI reduction (ERM IBMI=0.025) but over-suppresses feature variance along the spurious direction, inadvertently removing discriminative information and hurting WGA.

\begin{table}[!htbp]
\centering
\caption{BFR mode ablation on CelebA-Std R18, seed=1, $\lambda{=}20$.
  WGA (\%). \textbf{Bold}: best per method.}
\label{tab:mode_ablation}
\small
\begin{tabular}{lccccc}
\toprule
Mode & ERM & LfF & JTT & DFR & DebiAN \\
\midrule
Baseline (no BFR)     & 38.3 & 50.9 & 75.7 & 79.6 & 66.5 \\
\midrule
\texttt{orth}         & 37.8 & 62.8 & 76.9 & 78.3 & 66.1 \\
\texttt{spectral}     & 39.5 & 40.0 & 0.6 & 38.3 & 46.7 \\
\texttt{inv}          & \textbf{61.1} & \textbf{82.8} & \textbf{78.8} & \textbf{81.1} & \textbf{77.2} \\
\bottomrule
\end{tabular}
\end{table}

\section{\texorpdfstring{$\lambda$}{lambda} Sensitivity for BFR-inv}
\label{app:bfr_lambda}

\colortext{blue}{The one-seed mode ablation favors \texttt{inv} for the tested CelebA-Std setup. Table~\ref{tab:lambda} varies $\lambda$ for this mode. ERM improves through $\lambda=30$, while LfF, DFR, and DebiAN peak at 20 in this sweep. JTT is sensitive to stronger penalties: at $\lambda=30$, the accuracy-selected checkpoint has zero WGA, while a WGA-selected checkpoint reaches $73.9\%$. This is a checkpoint-selection issue as well as a regularization sensitivity result; the experiment does not isolate a gradient-conflict mechanism. We retain the empirical defaults of 20 for most methods and 1 for JTT, including the near-neutral Waterbirds comparison ($80.9\pm3.8\%$ versus $80.8\pm3.8\%$). These settings are not a universally validated tuning rule.}

\begin{table}[!htbp]
\centering
\caption{$\lambda$ sensitivity for BFR-\texttt{inv} on CelebA-Std R18,
  seed=1. WGA (\%), accuracy-based model selection.
  \textbf{Bold}: selected value per method.
  $^*$Accuracy-selected checkpoint has WGA$=$0; WGA-selected checkpoint achieves $73.9$.}
\label{tab:lambda}
\small
\begin{tabular}{lccccl}
\toprule
Method & $\lambda{=}5$ & $\lambda{=}10$ & $\lambda{=}20$ & $\lambda{=}30$ & Trend \\
\midrule
ERM    & 48.9 & 55.6 & \textbf{61.1} & 64.4 & $\uparrow$ monotone \\
LfF    & 82.2 & 82.2 & \textbf{82.8} & 80.0 & peaked at 20 \\
JTT    & \textbf{78.8} & 78.6 & 76.1 & $0.0^*$ & $\downarrow$, collapses at 30 \\
DFR    & 75.6 & 79.4 & \textbf{81.1} & 77.2 & peaked at 20 \\
DebiAN & 66.1 & 71.7 & \textbf{77.2} & 46.7 & peaked at 20 \\
\midrule
\multicolumn{6}{l}{\textit{Selected $\lambda$:} 1 for JTT; 20 for all others.} \\
\bottomrule
\end{tabular}
\end{table}

\section{Continued-Training Durability}
\label{app:exp1_finetune}

We further evaluate whether the WGA obtained by each method remains useful after ordinary training pressure is resumed. Starting from each converged checkpoint, we continue standard ERM training for 30 epochs on the original biased CelebA-Std training set and the original Blond classification task. This protocol is a durability test for the learned representation: if a method mainly changes the final decision rule while leaving shortcut structure available in the backbone, continued ERM training can again favor that structure.

Table~\ref{tab:exp1_finetune} reports WGA at the starting checkpoint and after 30 epochs. BFR gives higher checkpoint WGA for every base method and higher final WGA for five of the six methods. The clearest case is DFR, where the final WGA after continued training is $63.5\%$ with BFR compared with $43.7\%$ without BFR. BFR also yields higher final WGA for ERM, JTT, DebiAN, and GroupDRO, while BFR-LfF remains comparable but slightly lower than LfF.

\begin{figure}[!htbp]
\centering
\suppincludegraphics[width=0.92\textwidth]{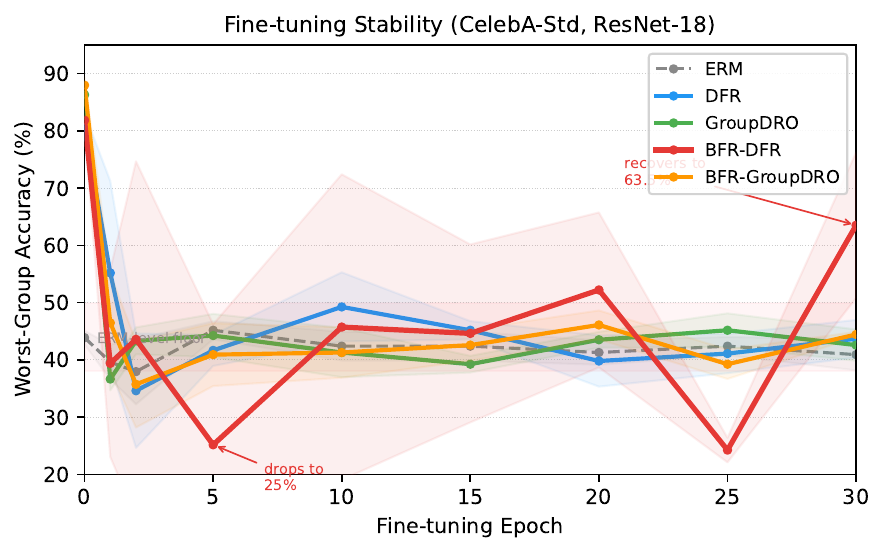}
\caption{WGA over 30 epochs of ERM continuation from each method's converged checkpoint on the original CelebA-Std Blond classification task. BFR-DFR reaches $63.5\%$ final WGA, substantially above DFR without BFR, indicating stronger retained WGA after continued training.}
\label{fig:exp1_trajectory}
\end{figure}

Figure~\ref{fig:exp1_trajectory} shows the full 30-epoch trajectory. We use this result as supporting evidence for durability rather than as the primary BFR claim: BFR most strongly improves retained WGA when paired with DFR, consistent with the view that geometric regularization is most useful when the downstream procedure would otherwise rely on a head-level correction over a contaminated backbone.

\begin{table}[!htbp]
\centering
\caption{Continued-training durability on CelebA-Std: WGA~(\%) after standard ERM continuation on the original Blond classification task and biased training distribution. Each metric is reported as base / +BFR. Bold marks the higher WGA within each pair.}
\label{tab:exp1_finetune}
\footnotesize
\setlength{\tabcolsep}{4pt}
\begin{tabular}{lcc}
\toprule
Method & Initial WGA (base / +BFR) & Final WGA (base / +BFR) \\
\midrule
ERM      & $43.9\pm1.2$ / \textbf{54.6$\pm$2.6} & $40.9\pm2.7$ / \textbf{42.6$\pm$1.1} \\
LfF      & $58.7\pm2.1$ / \textbf{61.7$\pm$2.3} & \textbf{46.9$\pm$0.3} / $46.1\pm2.1$ \\
JTT      & $81.1\pm0.4$ / \textbf{82.2$\pm$0.5} & $42.8\pm2.0$ / \textbf{44.8$\pm$2.1} \\
DFR      & $80.6\pm2.5$ / \textbf{81.9$\pm$1.0} & $43.7\pm3.3$ / \textbf{63.5$\pm$12.7} \\
DebiAN   & $64.4\pm3.2$ / \textbf{65.4$\pm$5.3} & $43.7\pm3.6$ / \textbf{45.7$\pm$2.5} \\
GroupDRO & $86.3\pm0.5$ / \textbf{88.0$\pm$1.5} & $42.6\pm2.7$ / \textbf{44.4$\pm$0.5} \\
\bottomrule
\end{tabular}
\end{table}

\section{Cross-Task Transfer Stress Test}
\label{app:exp3_transfer}

As an auxiliary stress test, we ask whether the backbone change reported in the main paper's frozen-head probing analysis carries over to a different CelebA task. This experiment is not intended as the primary evidence for BFR's backbone-level effect; the main evidence is the same-task frozen-head and Male-probe result. Here, we freeze the Blond/Male-trained backbone and train a fresh linear head on \emph{Smiling} prediction using the standard biased downstream training data.

This setting is deliberately outside the direct target of BFR. BFR suppresses the Male-aligned structure learned for Blond classification, whereas Smiling is a different target for which Male-correlated visual structure may be partially useful rather than purely spurious. Therefore, the appropriate interpretation is not that BFR should improve every cross-task metric. Instead, the experiment tests whether reducing the targeted spurious axis can improve downstream group balance when that axis is reused as a shortcut, and where this benefit stops when the new task uses related information.

Table~\ref{tab:exp3_transfer} therefore reports biased-head transfer metrics directly, rather than a difference between two WGA values from different head-training protocols. We include overall accuracy, mean group accuracy across the four Smiling/Male groups, WGA, and group disparity. This avoids making the conclusion depend on a gap between two worst-group statistics.

\colortext{blue}{BFR+GroupDRO improves the reported Smiling-transfer metrics: overall accuracy $80.4\%$ to $82.4\%$, mean group accuracy $79.8\%$ to $82.1\%$, WGA $74.0\%$ to $79.6\%$, and disparity $9.5\%$ to $5.5\%$. ERM also improves across these metrics, whereas DFR trades lower overall accuracy for better WGA. LfF, JTT, and DebiAN have lower transferred WGA with BFR. This mixed result defines the tested scope and does not establish a causal explanation for the gains or a universal transfer benefit.}

\begin{table}[!htbp]
\centering
\caption{Cross-task transfer to Smiling prediction with the Blond/Male backbone frozen and a fresh biased downstream head trained on standard Smiling data. Acc is overall accuracy; Mean Group Acc averages the four Smiling/Male groups; Group Disparity is the best-group minus worst-group accuracy. 3 seeds, mean$\pm$std. Each metric is reported as base / +BFR, and bold marks the better value within each pair.}
\label{tab:exp3_transfer}
\footnotesize
\setlength{\tabcolsep}{4pt}
\begin{tabular}{llcccc}
\toprule
Method & Paradigm & Acc ($\uparrow$) & Mean Group Acc ($\uparrow$) & WGA ($\uparrow$) & Group Disparity ($\downarrow$) \\
\midrule
ERM      & Baseline      & $81.1\pm0.5$ / \textbf{81.9$\pm$0.8} & $80.3\pm0.5$ / \textbf{81.2$\pm$0.8} & $74.2\pm0.5$ / \textbf{76.9$\pm$1.3} & $10.4\pm0.4$ / \textbf{6.5$\pm$0.5} \\
LfF      & Loss-rewt.    & \textbf{77.8$\pm$2.1} / $77.1\pm4.6$ & \textbf{77.1$\pm$2.1} / $76.5\pm4.6$ & \textbf{71.0$\pm$1.5} / $70.2\pm6.3$ & \textbf{10.9$\pm$0.2} / $11.9\pm2.4$ \\
JTT      & Loss-rewt.    & \textbf{80.1$\pm$0.4} / $79.4\pm0.3$ & \textbf{79.2$\pm$0.4} / $78.3\pm0.3$ & \textbf{71.3$\pm$0.5} / $68.0\pm0.4$ & \textbf{13.1$\pm$0.7} / $16.6\pm0.9$ \\
DFR      & Head-retrain  & \textbf{80.1$\pm$0.3} / $78.6\pm0.6$ & \textbf{79.3$\pm$0.4} / $78.1\pm0.6$ & $72.7\pm0.6$ / \textbf{73.5$\pm$1.1} & $10.8\pm0.2$ / \textbf{7.1$\pm$1.6} \\
DebiAN   & Loss-rewt.    & \textbf{80.2$\pm$2.1} / $75.7\pm5.7$ & \textbf{79.4$\pm$2.2} / $75.2\pm5.8$ & \textbf{72.9$\pm$2.6} / $69.5\pm9.7$ & $11.3\pm0.8$ / \textbf{10.9$\pm$6.4} \\
GroupDRO & Supervised    & $80.4\pm0.4$ / \textbf{82.4$\pm$0.3} & $79.8\pm0.4$ / \textbf{82.1$\pm$0.4} & $74.0\pm0.7$ / \textbf{79.6$\pm$0.3} & $9.5\pm0.3$ / \textbf{5.5$\pm$0.4} \\
\bottomrule
\end{tabular}
\end{table}

\section{Attribute Selectivity Analysis}
\label{app:exp4_surgical}

\textbf{Setup.} All experiments in this subsection use models trained on \textbf{CelebA-Std} for the \textbf{Blond Hair} classification task, with \textbf{Male} as the spurious attribute that BFR is instructed to suppress. After training, we \emph{freeze} the backbone and train lightweight linear probes on the frozen features to predict a range of \emph{other} CelebA attributes. These attributes were never part of the training objective and were not the spurious axis BFR was instructed to suppress. This design isolates the question: does BFR primarily suppress the Male representational axis, or does it degrade the backbone's encoding of unrelated attributes as well? If BFR is targeted, probes for attributes with low correlation to Male ($|r|{\approx}0$) should be unaffected, while probes for attributes that share geometric structure with Male ($|r|{\to}1$) should degrade proportionally, since reducing the Male axis can also reduce whatever signal those attributes were borrowing from it.

\colortext{blue}{A centroid penalty may change information useful for other tasks. We examine alternative groupings and probes for attributes with different label correlations to Male. These correlations are observational descriptors and do not identify orthogonal feature subspaces.}

\colortext{blue}{\textbf{Alternative evaluation groups.} On the original Blond task, the recorded BFR-DFR and BFR-GroupDRO comparisons are within approximately $1$\,pp under Young, Heavy\_Makeup, Attractive, and Wearing\_Lipstick grouping. This tests the existing predictor under alternative group definitions; it does not show that those attributes remain equally encoded. Separate attribute probes in Table~\ref{tab:exp4_spectrum} show losses for several of these attributes.}

\textbf{Attribute correlation spectrum.} We train linear probes for 12 CelebA attributes on the frozen backbone, spanning $|r|{=}0.013$ (Narrow\_Eyes, geometrically independent of Male) to $|r|{=}1.000$ (Male itself, the spurious axis BFR regularizes against), and report each method's per-attribute probe accuracy for both the Base and the +BFR backbone (Table~\ref{tab:exp4_spectrum}).

\begin{landscape}
\begin{table}[p]
	\centering
	\caption{Attribute correlation spectrum on CelebA-Std with ResNet-18, averaged over 3 seeds. Backbones are trained for \emph{Blond Hair} classification with \emph{Male} as the BFR-targeted spurious attribute, then frozen for linear-probe evaluation across 12 CelebA attributes sorted by $|r(\text{attr},\,\text{Male})|$. Each method reports Base (no BFR) / +BFR / $\Delta$ (BFR$-$Base). The bottom row reports Spearman $\rho$ between $|r|$ and probe accuracy; \textbf{bold} marks the more selective Base/+BFR value within each method.}
	\label{tab:exp4_spectrum}
	
	\scriptsize
	\setlength{\tabcolsep}{2.2pt}
	\resizebox{\linewidth}{!}{%
\begin{tabular}{lr ccc ccc ccc ccc ccc ccc}
		\toprule
		& & \multicolumn{3}{c}{\textbf{ERM}} & \multicolumn{3}{c}{\textbf{LfF}} & \multicolumn{3}{c}{\textbf{JTT}} & \multicolumn{3}{c}{\textbf{DFR}} & \multicolumn{3}{c}{\textbf{DebiAN}} & \multicolumn{3}{c}{\textbf{GroupDRO}} \\
		\cmidrule(lr){3-5}\cmidrule(lr){6-8}\cmidrule(lr){9-11}\cmidrule(lr){12-14}\cmidrule(lr){15-17}\cmidrule(lr){18-20}
		Attribute & $|r|$ & Base & +BFR & $\Delta$ & Base & +BFR & $\Delta$ & Base & +BFR & $\Delta$ & Base & +BFR & $\Delta$ & Base & +BFR & $\Delta$ & Base & +BFR & $\Delta$ \\
		\midrule
		Narrow\_Eyes      & .013 & 85.3 & 85.3 & $-0.0$ & 85.1 & 85.1 & $+0.0$ & 85.2 & 85.1 & $-0.1$ & 85.1 & 85.1 & $+0.0$ & 85.1 & 85.1 & $-0.0$ & 85.3 & 85.1 & $-0.2$ \\
		Blurry            & .026 & 95.5 & 95.1 & $-0.4$ & 95.2 & 94.9 & $-0.3$ & 95.6 & 95.0 & $-0.5$ & 95.4 & 94.9 & $-0.4$ & 95.3 & 95.0 & $-0.3$ & 95.5 & 94.9 & $-0.5$ \\
		Straight\_Hair    & .063 & 81.2 & 79.7 & $-1.5$ & 79.9 & 79.0 & $-0.9$ & 81.5 & 79.5 & $-2.0$ & 80.3 & 79.0 & $-1.2$ & 80.1 & 79.0 & $-1.1$ & 81.3 & 79.4 & $-1.9$ \\
		Pale\_Skin        & .076 & 96.0 & 95.8 & $-0.2$ & 95.9 & 95.8 & $-0.1$ & 96.0 & 95.8 & $-0.2$ & 95.9 & 95.8 & $-0.1$ & 96.0 & 95.8 & $-0.2$ & 96.0 & 95.8 & $-0.2$ \\
		Mouth\_Open       & .100 & 74.7 & 75.4 & $+0.7$ & 68.2 & 62.9 & $-5.3$ & 74.3 & 69.6 & $-4.7$ & 73.2 & 65.7 & $-7.4$ & 74.3 & 61.2 & $-13.2$ & 74.2 & 73.9 & $-0.3$ \\
		Brown\_Hair       & .109 & 85.7 & 85.1 & $-0.7$ & 85.6 & 82.2 & $-3.4$ & 86.4 & 84.8 & $-1.5$ & 84.8 & 82.2 & $-2.5$ & 86.0 & 82.7 & $-3.3$ & 85.8 & 85.5 & $-0.3$ \\
		Black\_Hair       & .111 & 85.5 & 82.8 & $-2.8$ & 82.9 & 74.8 & $-8.1$ & 86.1 & 82.0 & $-4.2$ & 83.0 & 73.1 & $-9.9$ & 85.0 & 77.2 & $-7.8$ & 85.9 & 82.5 & $-3.4$ \\
		Attractive        & .302 & 78.1 & 72.6 & $-5.6$ & 77.1 & 64.4 & $-12.7$ & 78.0 & 73.4 & $-4.6$ & 77.2 & 64.0 & $-13.2$ & 77.7 & 65.3 & $-12.4$ & 77.6 & 68.7 & $-8.9$ \\
		Young             & .440 & 83.9 & 81.2 & $-2.7$ & 82.6 & 76.4 & $-6.3$ & 83.2 & 80.2 & $-3.0$ & 83.1 & 75.9 & $-7.2$ & 83.5 & 77.6 & $-5.9$ & 83.7 & 80.1 & $-3.6$ \\
		Heavy\_Makeup     & .681 & 86.5 & 76.7 & $-9.8$ & 84.3 & 63.8 & $-20.5$ & 86.4 & 79.1 & $-7.3$ & 85.3 & 64.5 & $-20.8$ & 86.2 & 65.5 & $-20.7$ & 85.5 & 70.2 & $-15.3$ \\
		Wearing\_Lipstick & .789 & 89.2 & 76.6 & $-12.5$ & 87.6 & 58.0 & $-29.6$ & 89.3 & 81.0 & $-8.2$ & 87.6 & 59.4 & $-28.2$ & 89.2 & 61.8 & $-27.5$ & 88.4 & 67.8 & $-20.6$ \\
		Male (spurious)   & 1.00 & 93.2 & 80.0 & $-13.2$ & 90.6 & 61.4 & $-29.2$ & 93.4 & 86.6 & $-6.9$ & 91.9 & 67.4 & $-24.5$ & 92.8 & 65.2 & $-27.6$ & 92.5 & 72.2 & $-20.3$ \\
		\midrule
		Spearman $\rho(|r|,\text{acc})$ & & $+0.12$ & $\mathbf{-0.53}$ & $-0.65$ & $-0.01$ & $\mathbf{-0.80}$ & $-0.79$ & $+0.11$ & $\mathbf{-0.26}$ & $-0.37$ & $+0.06$ & $\mathbf{-0.75}$ & $-0.81$ & $+0.08$ & $\mathbf{-0.69}$ & $-0.77$ & $+0.08$ & $\mathbf{-0.71}$ & $-0.79$ \\
		\bottomrule
	\end{tabular}
}%

\end{table}
\end{landscape}

\colortext{blue}{Male probe accuracy measures recoverability under the fitted linear probe, not use by the original Blond classifier. BFR lowers this accuracy but does not make Male linearly unrecoverable in the reported tests. The study therefore evaluates partial suppression and retained utility, not perfect erasure. Low label correlation with Male also does not establish an orthogonal feature subspace or equal task difficulty.}

\colortext{blue}{We retain Spearman correlations between attribute--Male label correlation and raw attribute-probe accuracy as descriptive summaries. They shift from $[-0.01,+0.12]$ for base backbones to $[-0.80,-0.26]$ with BFR. These raw accuracies depend on attribute prevalence and task difficulty, so the rank correlations alone do not prove selective feature removal or reveal a geometrically isolated Male axis. The paired per-attribute changes are more informative about the limits of the intervention.}

\colortext{blue}{BFR-JTT has the smallest recorded correlation change ($\Delta\rho=-0.37$), while several stronger variants have larger shifts. The table also shows important collateral losses on weakly Male-correlated tasks, including Mouth\_Open (up to $13.2$\,pp lower) and Black\_Hair (up to $9.9$\,pp lower). These negative results restrict claims of information preservation. Averaging or rank summaries do not eliminate those losses.}

\begin{figure}[!htbp]
\centering
\suppincludegraphics[width=0.92\textwidth]{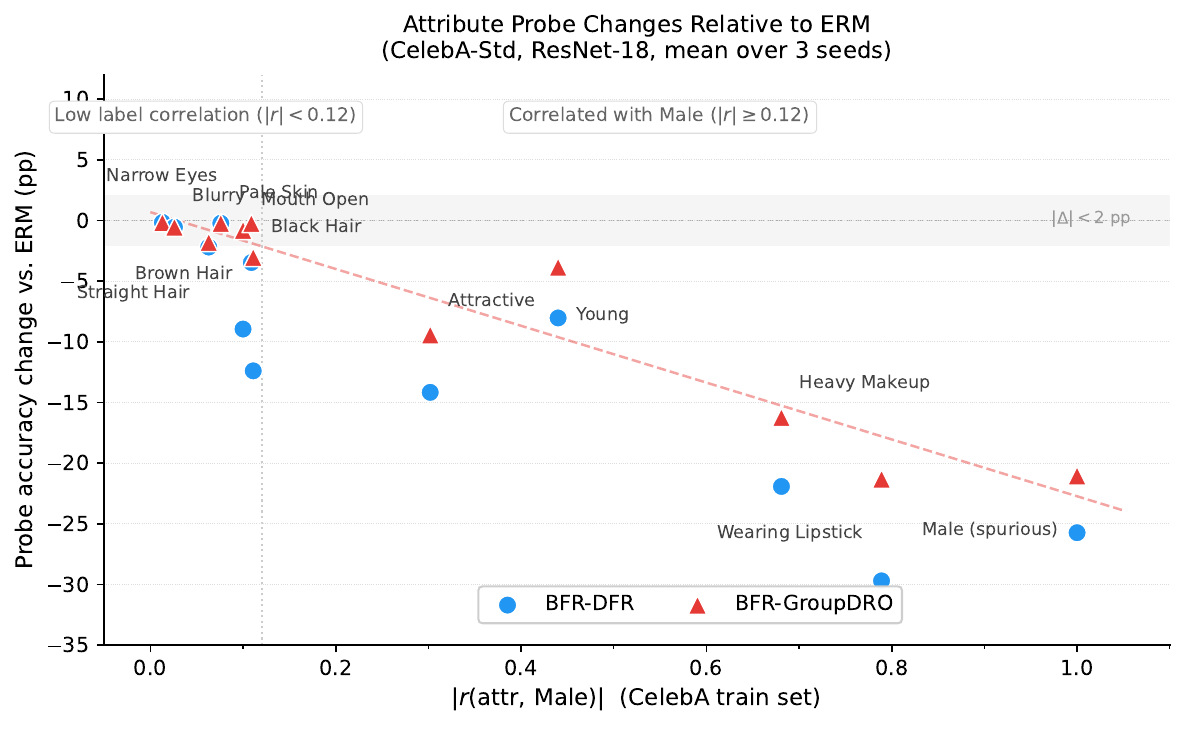}
\caption{\colortext{blue}{Attribute-probe accuracy differences relative to ERM vs.\ absolute label correlation with Male, for BFR-DFR and BFR-GroupDRO. The saved plotting source uses ERM as the reference for both curves; this is not the paired BFR-minus-corresponding-base comparison in Table~\ref{tab:exp4_spectrum}. Larger losses occur for several Male-correlated attributes, with exceptions and collateral losses at low correlation. The plot does not isolate the within-method effect of BFR or establish a distinct erased subspace.}}
\label{fig:exp4_gradient}
\end{figure}

\clearpage
\section{Frozen-Backbone Probe Details}
\label{app:probe_details}
\colortext{blue}{This table reconstructs means and sample standard deviations from the saved per-seed JSON outputs (seeds 1, 245, 392) for the experiments already reported in the main paper. It adds no new training or evaluations. All backbones remain fixed. The task heads use SGD on biased training features or L-BFGS on balanced validation features; the Male probe uses SGD on validation features and test evaluation. Separate backbone paths are recorded for base DFR and ERM, so these are not a shared-checkpoint identity test. Raw attribute accuracy is not corrected for label prevalence. Three seeds provide limited uncertainty characterization and no significance or equivalence test is claimed.}
\begin{table}[!htbp]
\centering
\caption{\colortext{blue}{Existing frozen-backbone probe results, mean $\pm$ sample standard deviation over three seeds, percentages. Each pair is base / +BFR. Balanced WGA is included as a protocol check, not an optimal-head bound.}}
\label{tab:probe_uncertainty}
\small
\begin{tabular}{lccc}
\toprule
Method & Biased test WGA & Male test accuracy & Balanced test WGA \\
\midrule
\colortext{blue}{ERM} & \colortext{blue}{$40.0\pm1.5$ / $44.6\pm5.3$} & \colortext{blue}{$93.2\pm0.1$ / $80.0\pm0.1$} & \colortext{blue}{$83.3\pm2.9$ / $81.3\pm5.0$ } \\
\colortext{blue}{LfF} & \colortext{blue}{$44.8\pm5.2$ / $60.0\pm3.5$} & \colortext{blue}{$90.6\pm2.5$ / $61.4\pm7.1$} & \colortext{blue}{$80.2\pm4.5$ / $82.6\pm2.6$ } \\
\colortext{blue}{JTT} & \colortext{blue}{$36.5\pm2.2$ / $42.2\pm0.6$} & \colortext{blue}{$93.4\pm0.3$ / $86.6\pm0.4$} & \colortext{blue}{$81.3\pm1.3$ / $79.6\pm1.8$ } \\
\colortext{blue}{DFR} & \colortext{blue}{$42.2\pm1.0$ / $44.4\pm3.6$} & \colortext{blue}{$91.9\pm0.2$ / $67.5\pm1.1$} & \colortext{blue}{$80.2\pm3.9$ / $79.8\pm0.6$ } \\
\colortext{blue}{DebiAN} & \colortext{blue}{$45.4\pm0.6$ / $52.6\pm2.7$} & \colortext{blue}{$92.8\pm1.1$ / $65.2\pm8.9$} & \colortext{blue}{$80.0\pm2.9$ / $79.6\pm0.8$ } \\
\colortext{blue}{GroupDRO} & \colortext{blue}{$40.7\pm2.5$ / $64.1\pm3.5$} & \colortext{blue}{$92.5\pm0.1$ / $72.2\pm2.7$} & \colortext{blue}{$81.3\pm2.2$ / $84.4\pm6.2$ } \\
\bottomrule
\end{tabular}
\end{table}